\documentclass[journal,transmag]{IEEEtran}
\ifCLASSINFOpdf
\else
\fi
\usepackage{url}
\usepackage{times}
\usepackage{epsfig}
\usepackage{graphicx}
\usepackage{amsmath}
\usepackage{amssymb}
\usepackage{subfigure}
\usepackage{upgreek}
\usepackage{multirow}
\usepackage{color}
\usepackage{bm}

\usepackage{arydshln}
\usepackage{cite}
\usepackage{pdfpages}
\usepackage{blindtext}

\usepackage{amsthm,xpatch}

\xpatchcmd{\proof}{\hskip\labelsep}{\hskip3\labelsep}{}{}

\usepackage{caption}
\usepackage{xcolor}

\usepackage[pagebackref=true,breaklinks=true,letterpaper=true,colorlinks,bookmarks=false]{hyperref}

\begin{document}
%
% paper title
% Titles are generally capitalized except for words such as a, an, and, as,
% at, but, by, for, in, nor, of, on, or, the, to and up, which are usually
% not capitalized unless they are the first or last word of the title.
% Linebreaks \\ can be used within to get better formatting as desired.
% Do not put math or special symbols in the title.

\title{Real-world Noisy Image Denoising: A New Benchmark}

% author names and affiliations
% transmag papers use the long conference author name format.

\author{
\IEEEauthorblockN{Jun Xu$^{1}$,
Hui Li$^{1}$,
Zhetong Liang$^{1}$,
David Zhang$^{1,2}$,~\IEEEmembership{Fellow,~IEEE},
and
Lei Zhang\textsuperscript{1,*},~\IEEEmembership{Fellow,~IEEE}
}
\IEEEauthorblockA{$^{1}$Department of Computing,
The Hong Kong Polytechnic University, Hong Kong SAR, China
\\
$^{2}$School of Science and Engineering, The Chinese University of Hong Kong (Shenzhen), Shenzhen, China
}

%\thanks{Manuscript received April 1, 2017. 
%Corresponding author: Lei Zhang (email: cslzhang@polyu.edu.hk).}

}

% The paper headers
\markboth{}%
{Shell \MakeLowercase{\textit{et al.}}: Bare Demo of IEEEtran.cls for IEEE Transactions on Magnetics Journals}
% The only time the second header will appear is for the odd numbered pages
% after the title page when using the twoside option.
% 
% *** Note that you probably will NOT want to include the author's ***
% *** name in the headers of peer review papers.                   ***
% You can use \ifCLASSOPTIONpeerreview for conditional compilation here if
% you desire.

% If you want to put a publisher's ID mark on the page you can do it like
% this:
%\IEEEpubid{0000--0000/00\$00.00~\copyright~2015 IEEE}
% Remember, if you use this you must call \IEEEpubidadjcol in the second
% column for its text to clear the IEEEpubid mark.

% use for special paper notices
%\IEEEspecialpapernotice{(Invited Paper)}

% for Transactions on Magnetics papers, we must declare the abstract and
% index terms PRIOR to the title within the \IEEEtitleabstractindextext
% IEEEtran command as these need to go into the title area created by
% \maketitle.
% As a general rule, do not put math, special symbols or citations
% in the abstract or keywords.
\IEEEtitleabstractindextext{%
\begin{abstract}
Abstract: Most of previous image denoising methods focus on additive white Gaussian noise (AWGN). However,the real-world noisy image denoising problem with the advancing of the computer vision techiniques. In order to promote the study on this problem while implementing the concurrent real-world image denoising datasets, we construct a new benchmark dataset which contains comprehensive real-world noisy images of different natural scenes. These images are captured by different cameras under different camera settings. We evaluate the different denoising methods on our new dataset as well as previous datasets. Extensive experimental results demonstrate that the recently proposed methods designed specifically for realistic noise removal based on sparse or low rank theories achieve better denoising performance and are more robust than other competing methods, and the newly proposed dataset is more challenging. The constructed dataset of real photographs is publicly available at \url{https://github.com/csjunxu/PolyUDataset} for researchers to investigate new real-world image denoising methods. We will add more analysis on the noise statistics in the  real photographs of our new dataset in the next version of this article.
\end{abstract}

% Note that keywords are not normally used for peerreview papers.
\begin{IEEEkeywords}
Image denoising, Real-world noisy images, benchmark datasets
\end{IEEEkeywords}}

% make the title area
\maketitle

% To allow for easy dual compilation without having to reenter the
% abstract/keywords data, the \IEEEtitleabstractindextext text will
% not be used in maketitle, but will appear (i.e., to be "transported")
% here as \IEEEdisplaynontitleabstractindextext when the compsoc 
% or transmag modes are not selected <OR> if conference mode is selected 
% - because all conference papers position the abstract like regular
% papers do.
\IEEEdisplaynontitleabstractindextext
% \IEEEdisplaynontitleabstractindextext has no effect when using
% compsoc or transmag under a non-conference mode.

% For peer review papers, you can put extra information on the cover
% page as needed:
% \ifCLASSOPTIONpeerreview
% \begin{center} \bfseries EDICS Category: 3-BBND \end{center}
% \fi
%
% For peerreview papers, this IEEEtran command inserts a page break and
% creates the second title. It will be ignored for other modes.
\IEEEpeerreviewmaketitle

\vspace{-1mm}
\section{Introduction}
%\vspace{-1mm}

\IEEEPARstart{D}{uring} the past decades, the statistical property of real-world noise has been studied for CCD and CMOS image sensors \cite{healey1994radiometric,tsin2001statistical,RENOIR2014,crosschannel2016,dnd2017}. There are five major sources for real-world noise, including photon shot noise, fixed pattern noise, dark current, readout noise, and quantization noise, etc. The shot noise is one inevitable source of noise, which is induced by the stochastic arrival process of photons to the sensor. The arrival of photons can be modeled by a Possion process in which the number of photons arriving the sensor follows a Possion distribution. This type of noise is proportional to the mean intensity of the specific pixel and is not stationary across the whole image. The fixed pattern noise include pixel response non-uniformity (PRNU) noise and dark current non-uniformity (DCNU) noise. In PRNU noise, each pixel will have a slightly different output level or response for a fixed light level. The major cause of the PRNU noise is the loss of light and color mixture in the neighboring pixels. The DCNU noise comes from the electronics within the sensor chip, and it is generated due to thermal agitation, even there is no light reaching the camera sensor. The readout noise and quantization noise come from the discretization of measured signals. The readout noise is generated during the process of charge-to-voltage conversion, which is inheretantly not accurate. The quantization noise is generated when the readout values are quantized to integers. The final pixel values are discretizations of the original raw pixel values. Other noise include CCD specific sources such as transfer efficiency, and CMOS specific sources such as column noise.

Different from additive white Gaussian noise (AWGN), the real-world noise is signal dependent, and cannot be modeled by an explicit distribution. It becomes much more complex after being processed in the camera imaging pipelines. Hence, removing noise from real-world noisy images is a more challenging task than its synthetic AWGN counterpart. Another issue about real-world image denoising is how to evaluate the quality of the denoised images. The image quality assessment by subjective evaluation would be time-consuming, since it needs huge number of subjectives to take part in the evaluation experiments. An alternative is to resort to the objective evaluation. However, since the real-world noisy images have no corresponding ``ground truth'' images, the objective evaluation on the quality of denoised images is very hard. Another choice is to resort to some blind image quality assessment (BIQA) methods \cite{bliinds,biqi}. However, these BIQA methods are mostly developed based on the commonly used datasets such as TID dataset \cite{tid2008} and LIVE IQA dataset \cite{LIVEIQA}, whose images have very different properties from real-world noisy images.

Recently, several works have been done to address the issue of missing corresponding ``ground truth'' image of the captured real-world noisy image. In \cite{crosschannel2016}, a dataset containing 11 scenes is constructed for analyzing the properties of real-world noise produced in the camera imaging pipeline. However, this dataset is limited in several aspects. It contains only printed pictures on the package of several products, having few real objects. Other problems include that the intensity transform does not model heteroscedastic noise, and low-frequency bias is not removed, etc. In \cite{RENOIR2014} and \cite{dnd2017}, the corresponding ``ground truth'' image of the captured real-world noisy image is captured with low ISO values (e.g., ISO=100), with other post-processing steps such as linear intensity changes, spatial misalignment, and low-frequency residual correciton, etc. However, the ``ground truth'' images with low ISO values may have slightly different illuminations from the corresponding real-world noisy images captured under high ISO values (e.g., ISO=6,400). Besides, the post-processing steps may introduce human bias into the ``ground truth'' images. The work of \cite{EMVA1288} proposes a less tedious capture protocol similar to \cite{dnd2017}, where multiple exposures of a static scene are used to aggregate the measurements at every pixel site temporally. The works of \cite{noisemeasurement,moldovan2006denoising} propose to illuminate the sensor with approximately constant irradiation and subsequently aggregates intensity measurements spatially. This is repeated for different irradiation levels to capture the intensity dependence of the noise. In contrast, in \cite{dnd2017} the employed Tobit regression allows to estimate the parameters of the noise process by having access to just two images.

In this work, we construct a large dataset of real-world noisy images with reasonably obtained corresponding ``ground truth'' images. The basic idea is to capture the same and unchanged scene for many (e.g., 500) times and compute their mean image, which can be roughly taken as the ``ground truth'' image for the real-world noisy images. The rational of this strategy is that for each pixel, the noise is generated randomly larger or smaller than 0. Sampling the same pixel many times and computing the average value will approximate the truth pixel value and alleviate significantly the noise.

\section{Existing Datasets}

Currently, there are some datasets available aiming at benchmarking the denoising methods on real-world noisy images \cite{RENOIR2014,crosschannel2016,dnd2017}.

As far as we know, the RENOIR dataset \cite{RENOIR2014} is the first dataset on real-world noisy images with ``ground truth'' noise-free images. The cameras used in this dataset are Canon Rebel T3i, Canon S90, and Xiaomi T3i. The authors took photos of a static scene with different ISO values. However, the post-processing is less refined. Image pairs appear to exhibit spatial misalignment, the intensity transform does not model heteroscedastic noise, and low-frequency bias is not removed. In \cite{RENOIR2014}, experiments have been conducted to validate that ignoring these factors makes the dataset less usefull. It is often useful to measure the noise characteristics of a sensor at a certain ISO level. It was proposed \cite{RENOIR2014} to illuminate the sensor with approximately constant irradiation and subsequently aggregate intensity measurements spatially. This is repeated for different irradiation levels to capture the intensity dependence of the noise. A less tedious capture protocol was also proposed in \cite{RENOIR2014}, where multiple exposures of a static scene are used to aggregate the measurements at every pixel site temporally. The detailed description of this dataset is listed in Table \ref{tab6-1}.

\begin{table*}[t!]
\caption{Cameras and camera settings used in the dataset \cite{RENOIR2014}.}
\label{tab6-1}
\begin{center}
\small
\renewcommand\arraystretch{1.2}
\begin{tabular*}{1\textwidth}{@{\extracolsep{\fill}}c|cc|cc|cc}
\hline
\multirow{2}{*}{Camera}
&
\multirow{2}{*}{Sensor Size (mm)}
&
\multirow{2}{*}{\# of Scenes}
&
\multicolumn{2}{c|}{``Ground Truth''}
&
\multicolumn{2}{c}{Noisy Image}
\\
&
&
&
ISO
&
Time (s)
&
ISO
&
Time (s)
\\
\hline
Canon S90 & $7.4\times5.6$ & 40 & 100  & 3.2  & 640, 1k & Auto 
\\
\hline   
Canon T3i & $22.3\times14.9$ & 40 & 100 & Auto  & 3.2k, 6.4k & Auto
\\
\hline
Xiaomi Mi3 & $4.69\times3.52$ & 40 & 100  & Auto  & 1.6k, 3.2k & Auto
\\
\hline
\end{tabular*}
\end{center}
\end{table*}

The second work along this direction is reported in \cite{crosschannel2016}, which involves 11 static scenes. The real-world noisy images and the corresponding ``ground truth'' images are collected. For each scene, 500 JPEG images are captured and the mean image of the 500 images is roughly taken as the ``ground truth'' image. Utilizing the mean of temporal images as ``ground truth'' image has also been employed in \cite{Liu2008,liupractical}, but the authors did not build a benchmark dataset. In the dataset of \cite{crosschannel2016}, the images are mostly with resolution of $7630\times4912$ and captured by Nikon D800 (ISO=1,600, 3,200, and 6,400), Nikon D600 (ISO=3,200), and Canon 5D Mark III (ISO=3,200). There are totally 15 cropped regions of size $512\times512$ provided for evaluating different denoising methods. The major problems of this dataset is that the captured images are almost printed scenes, which share similar noise statistical property. The camera settings such as the ISO values are also somewhat limited. The detailed description of this dataset is listed in Table \ref{tab6-2}. 

\begin{table*}[t!]
\caption{The detailed information of the cropped regions from the dataset \cite{crosschannel2016}.}
\label{tab6-2}
\begin{center}
\small
\renewcommand\arraystretch{1.2}
\begin{tabular*}{1\textwidth}{@{\extracolsep{\fill}}ccccc}
\hline
Camera
& 
ISO
&
\# of Images
&
JPEG
&
Image Size
\\
\hline
Canon 5D Mark III & 3.2k  & 3  & Fine & $512\times512$
\\
\hline
Nikon D600 & 3.2k & 3  & Normal & $512\times512$
\\
\hline   
Nikon D800 & 1.6k, 3.2k, 6.4k & 9  & Normal & $512\times512$
\\
\hline
\end{tabular*}
\end{center}
\vspace{-4mm}
\end{table*}

One recent benchmark is reported in \cite{dnd2017}. Different from the previous two datasets in \cite{RENOIR2014} and \cite{crosschannel2016}, \cite{dnd2017} employs the Tobit regression to estimate the parameters of the noise process by accessing just two images. In order to obviate the unrealistic setting by developing a methodology for benchmarking denoising techniques on real photographs, the authors of \cite{dnd2017} captured 50 different pairs of images with different ISO settings and shutter speeds. The image captured with high ISO and faster shutter speed is taken as the real-world noisy image, while the image captured with low ISO and slower shutter speed is roughly taken as the ``ground truth'' image. To derive better ``ground truth'', careful post-processing is designed in \cite{dnd2017}. The authors corrected spatial misalignment, coped with the inaccurate exposure parameters through a linear intensity transform based on a novel heteroscedastic Tobit regression model, and removed residual low-frequency bias that stems from minor illumination changes, etc. The proposed dataset is called the Darmstadt Noise Dataset (DND) \cite{dnd2017}, in which the cameras used for capturing the dataset include Sony A7R, Olympus E-M10, Sony RX100 IV, and Huawei Nexus 6P. The authors extracted the linear raw intensities from the captured images using the free software \textsl{Dcraw}, and then normalized the image intensities to the range of $[0, 1]$ by scaling the black and white levels. One interesting finding is that various denoising techniques that perform well on synthetic noisy images are clearly outperformed by BM3D \cite{bm3d} on realistic photographs. This benchmark delineates realistic evaluation scenarios that deviate strongly from those commonly used in the scientific literature. The detailed description of this dataset is listed in Table \ref{tab6-3}.

\begin{table*}[t!]
\caption{Cameras and camera settings used in the dataset \cite{dnd2017}.}

\label{tab6-3}
\begin{center}
\small
\renewcommand\arraystretch{1.2}
\begin{tabular*}{1\textwidth}{@{\extracolsep{\fill}}cccc}
\hline
Camera
&
\# of Scenes
&
Sensor Size (mm)
&
ISO
\\
\hline
Sony A7R & 13  & $36\times24$  & 100-25.6k
\\
\hline
Olympus E-M10 & 13  & $17.3\times13$  & 200-25.6k 
\\
\hline   
Sony RX100 IV & 12 & $13.2\times8.8$  & 125-8k 
\\
\hline   
Huawei Nexus 6P & 12 & $6.17\times4.55$  & 100-6.4k 
\\
\hline
\end{tabular*}
\end{center}
\vspace{-4mm}
\end{table*}

\section{The Proposed Dataset}

\subsection{Motivation}
As discussed previously, existing real-world noisy image datasets \cite{RENOIR2014,crosschannel2016,dnd2017} have several limitations in evaluating existing and future image denoising methods. These limitations include camera brands, camera settings, and captured scenes, etc.

\textbf{Camera Brands}: In the RENOIR dataset \cite{RENOIR2014}, the authors used two different camera brands, Canon (T3i and S90) and Xiaomi (Mi3), for image collection. In the dataset \cite{crosschannel2016}, the authors also used two camera brands, i.e., the Canon (5D) and Nikon (D600 and D800), for image collection. In the DND dataset \cite{dnd2017}, the authors used three different cameras including Sony (A7R and RX100 IV), Olympus (E-M10), and Huawei (Nexus 6P).

\textbf{Camera Settings}: In the RENOIR dataset \cite{RENOIR2014}, the ``ground truth'' images are all captured by setting the ISO as 100. The ISO in noisy images are set as follows: for Xiaomi Mi3, the ISO is set as 1,600 or 3,200; for Canon S90, the ISO is set as 640 or 1,000; for Canon T3i, the ISO is set as 3,200 or 6,400. For all the cases except for the reference image of Canon S90, the shutter speed is set as automatic. For Canon S90, the shutter speed is set as 3.2 seconds. In the dataset \cite{crosschannel2016}, three different ISOs (e.g., 1,600, 3,200, and 6,400) are employed when capturing images with Nikon D800, while ISO=3,200 is utilized for Canon 5D and Nikon D600. The DND dataset \cite{dnd2017}, the ranges of ISO are $100\sim25,600$ for Sony A7R, $200\sim25,600$ for Olympus E-M10, $125\sim8,000$ for Sony RX100 IV, and $100\sim6,400$ for Huawei Nexus 6P, respectively. 

\textbf{Captured Scenes}: The RENOIR dataset \cite{RENOIR2014} captures 40 scenes for each camera brand, and overall 120 scenes are included in the dataset. However, the noisy images and corresponding ``ground truth'' images in this dataset have distinct color difference, which is largely caused by inconsistent lighting conditions. The dataset \cite{crosschannel2016} contains only 11 indoor scenes, which are overlapped by similar contents and objects. Though containing 50 different scenes, the ``ground truth'' images in the DND dataset \cite{dnd2017} are not accessible yet. This limits the evaluation of the proposed denoising methods on visual quality.

\textbf{Discussion}. Among the above mentioned factors, the camera settings are very important when we capture the real-world noisy images, while the camera brands and captured scenes are relatively easy to improve. The camera settings include mainly ISO value, the shutter speed, and the aperture, etc. In general, the faster the shutter speed, the darker the captured images when we fix the other settings, and vice versa. Similarly, the smaller the ISO value (or aperture), the darker the captured images when we fix the other settings, and vice versa. 

In order to make the image less affected by the change of environment (e.g., object motion, change of illumination, camera shake, etc.), the shutter speed should be set as faster as possible. For example, the shutter speed of the Sony A7II camera is between 1/80,000 second and 30 seconds. Given suitable aperture and ISO, it is possible to capture images with normal illuminations when we set the shutter speed between 1/100 second and 1 second. The aperture could be set as any value as long as it is in the reasonable range. The aperture of the Sony camera is between F3.5 and F22. Setting the aperture between F3.5 and F15 can allow us to obtain images with normal illumination under the fixed ranges of shutter and ISO. In our capturing process, we fixed the shutter and aperture in a suitable range, and tuned the ISO values according to the given camera. In general, the noise level would be higher when the ISO is higher. We set the ISO values from a low value to a high value with fixed gap to more comprehensively evaluate the denoising methods.

\begin{figure*}[t!]
    \centering
\subfigure{
\begin{minipage}[t]{0.32\textwidth}
\centering
\raisebox{-0.5cm}{\includegraphics[width=1\textwidth]{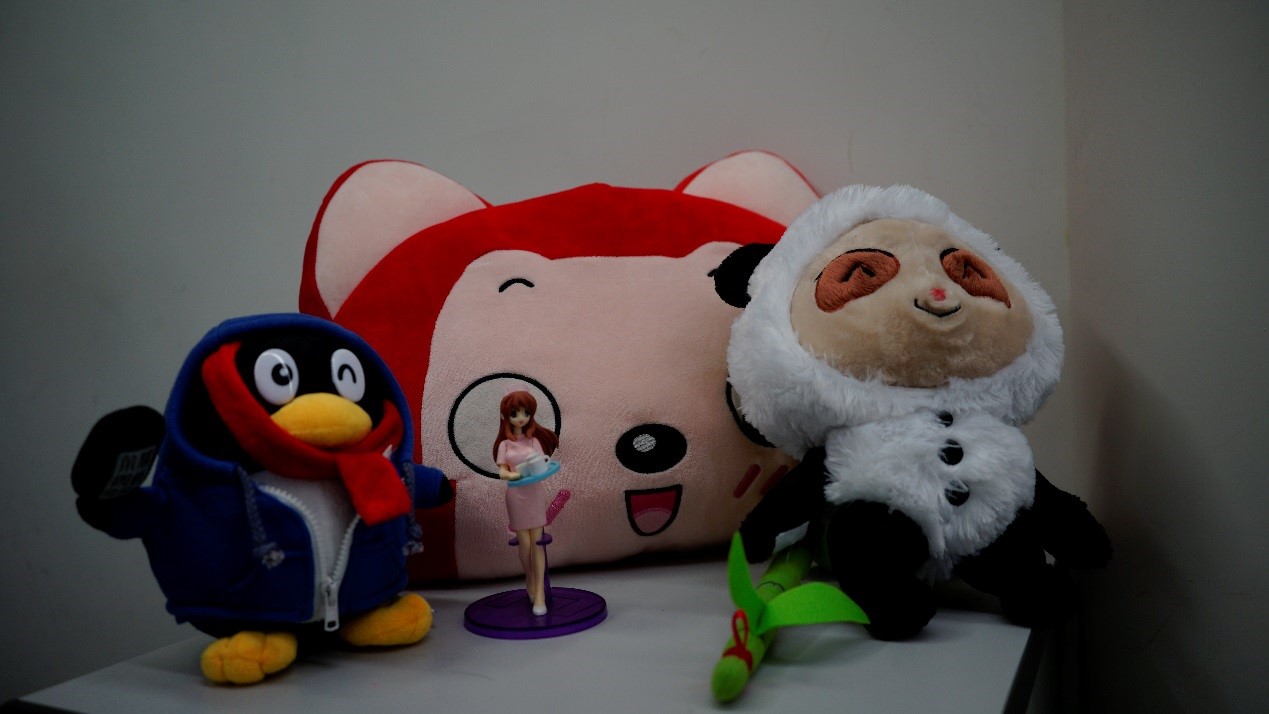}}
{\footnotesize (a) 200,3.5,1/60}
\end{minipage}
\begin{minipage}[t]{0.32\textwidth}
\centering
\raisebox{-0.5cm}{\includegraphics[width=1\textwidth]{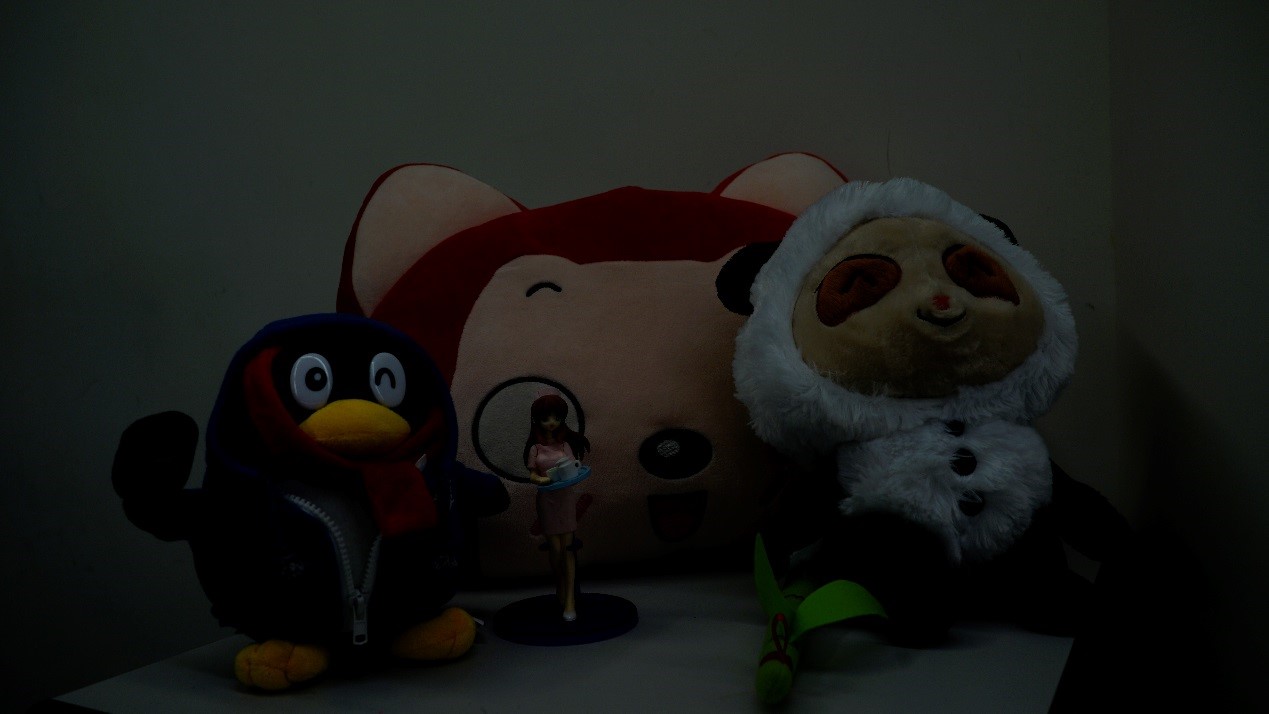}}
{\footnotesize (b) 200,6.7,1/60}
\end{minipage}
\begin{minipage}[t]{0.32\textwidth}
\centering
\raisebox{-0.5cm}{\includegraphics[width=1\textwidth]{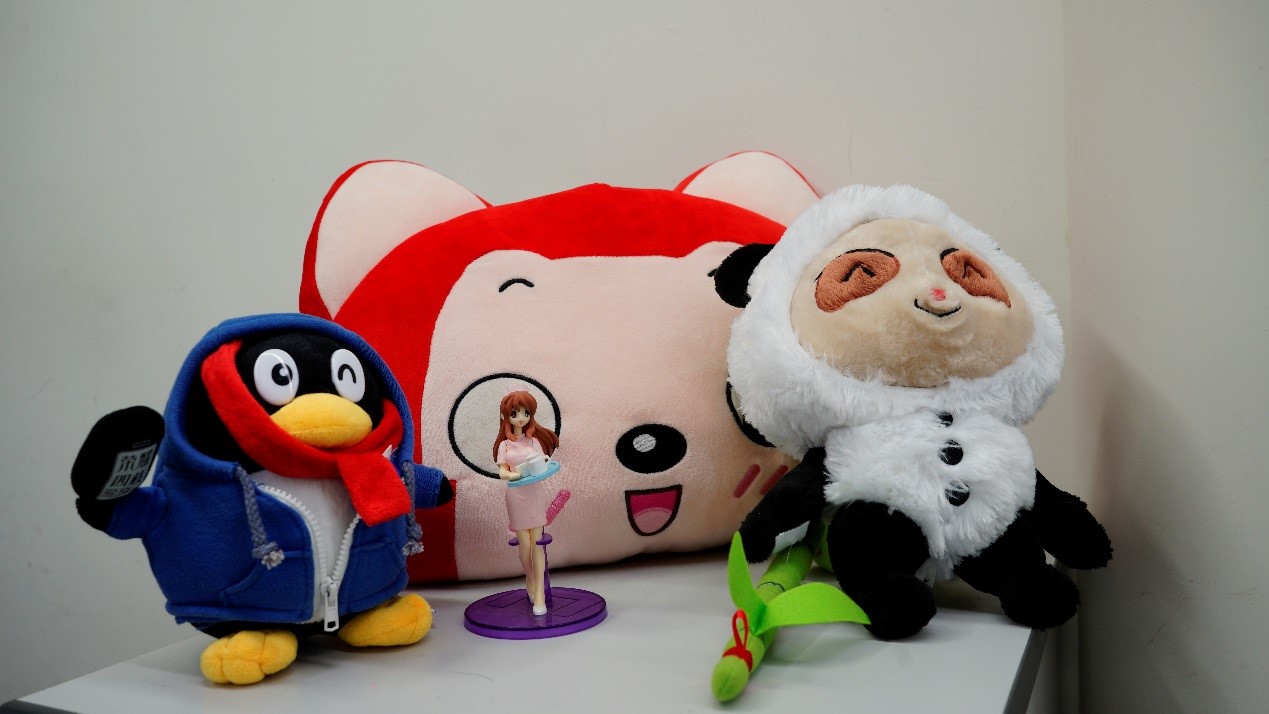}}
{\footnotesize (c) 200,6.7,1/8}
\end{minipage}
}\vspace{-3mm}
\subfigure{
\begin{minipage}[t]{0.32\textwidth}
\centering
\raisebox{-0.5cm}{\includegraphics[width=1\textwidth]{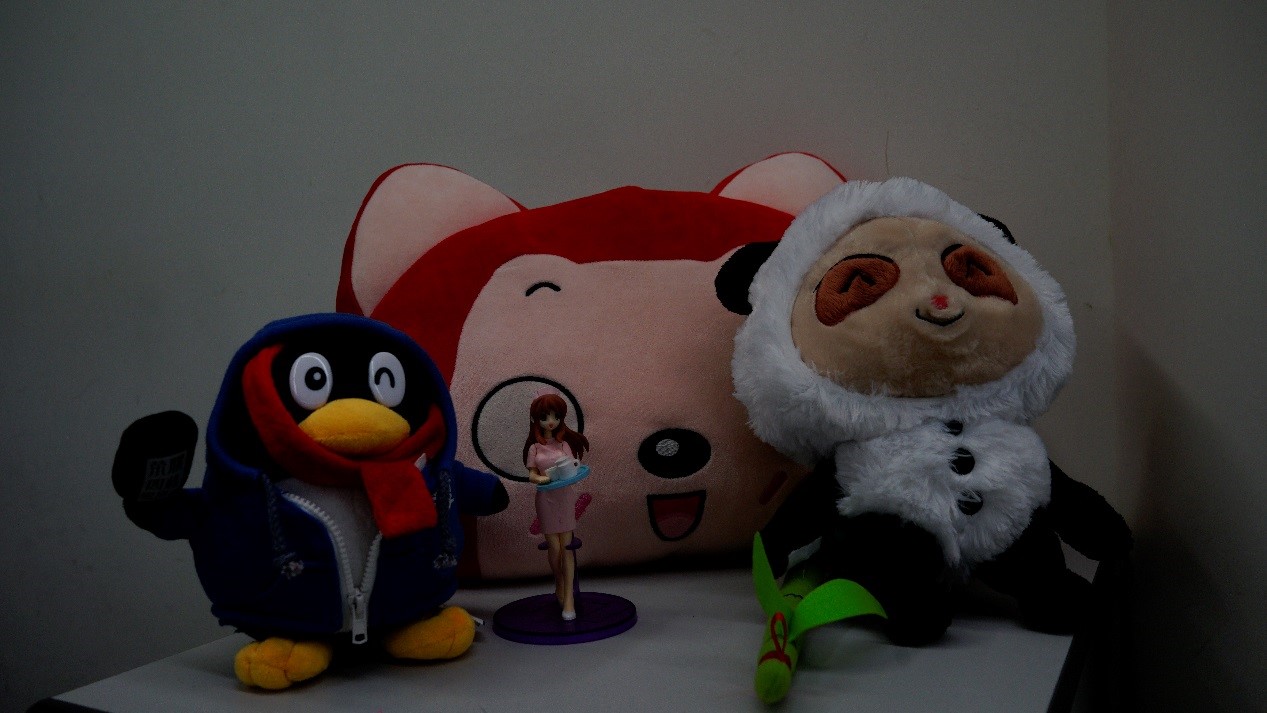}}
{\footnotesize (d) 400,6.7,1/60}
\end{minipage}
\begin{minipage}[t]{0.32\textwidth}
\centering
\raisebox{-0.5cm}{\includegraphics[width=1\textwidth]{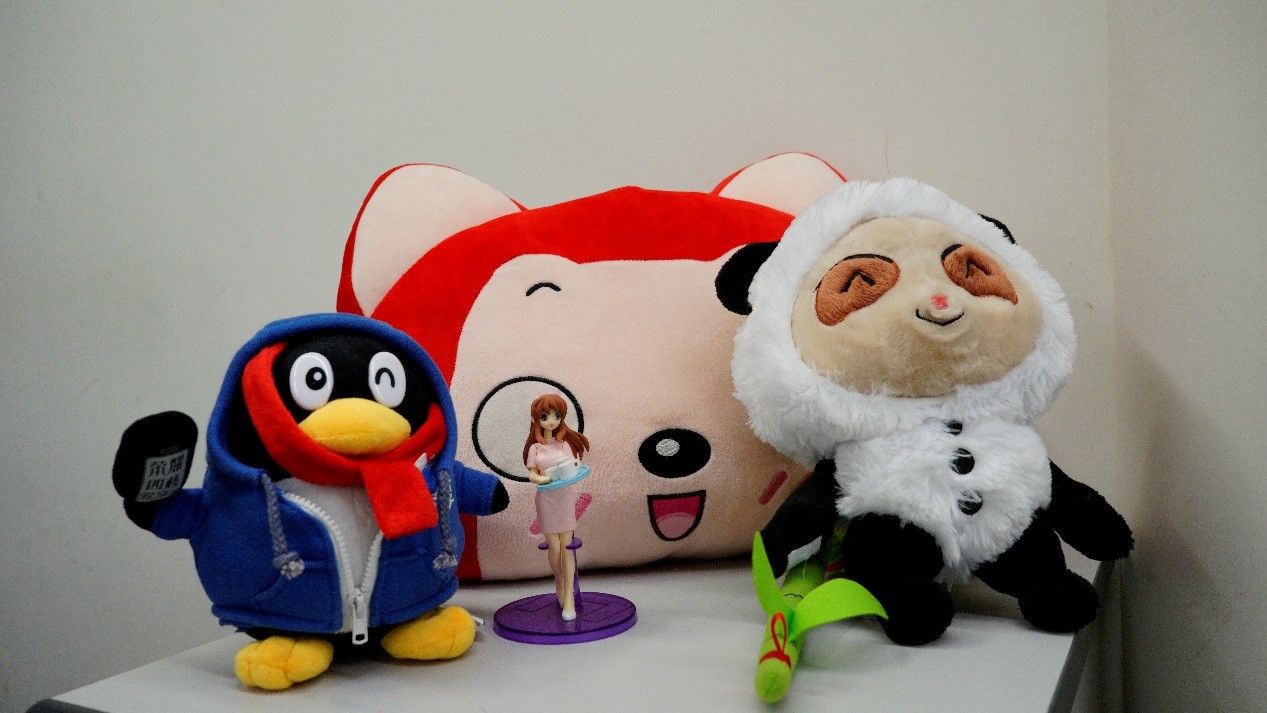}}
{\footnotesize (e) 1600,6.7,1/60}
\end{minipage}
\begin{minipage}[t]{0.32\textwidth}
\centering
\raisebox{-0.5cm}{\includegraphics[width=1\textwidth]{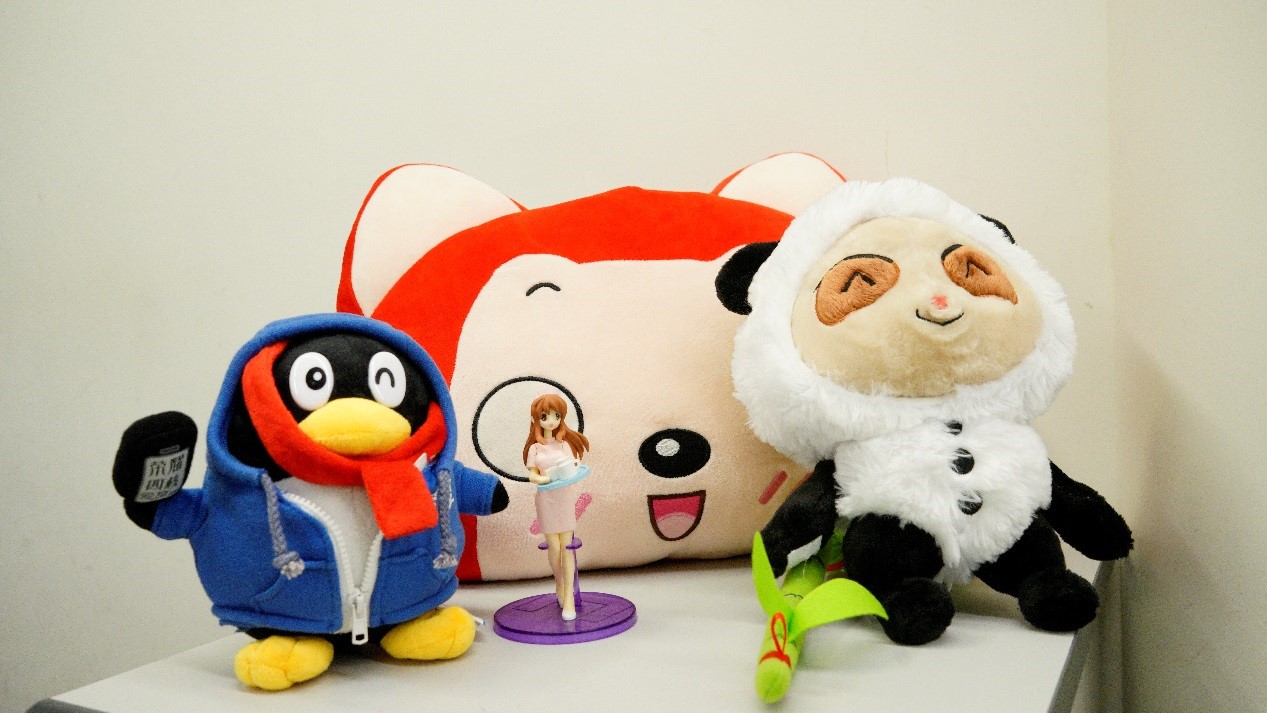}}
{\footnotesize (f) 3200,6.7,1/60}
\end{minipage}
}\vspace{-3mm}
\subfigure{
\begin{minipage}[t]{0.32\textwidth}
\centering
\raisebox{-0.5cm}{\includegraphics[width=1\textwidth]{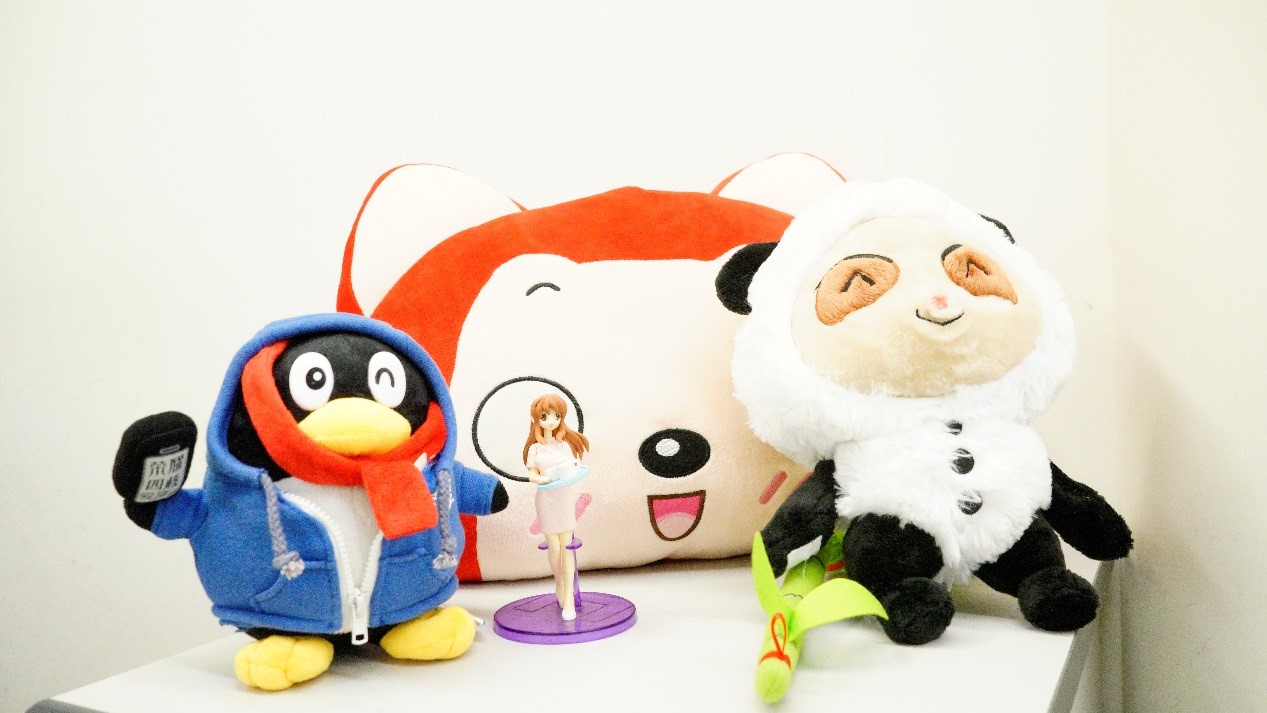}}
{\footnotesize (g) 6400,6.7,1/60}
\end{minipage}
\begin{minipage}[t]{0.32\textwidth}
\centering
\raisebox{-0.5cm}{\includegraphics[width=1\textwidth]{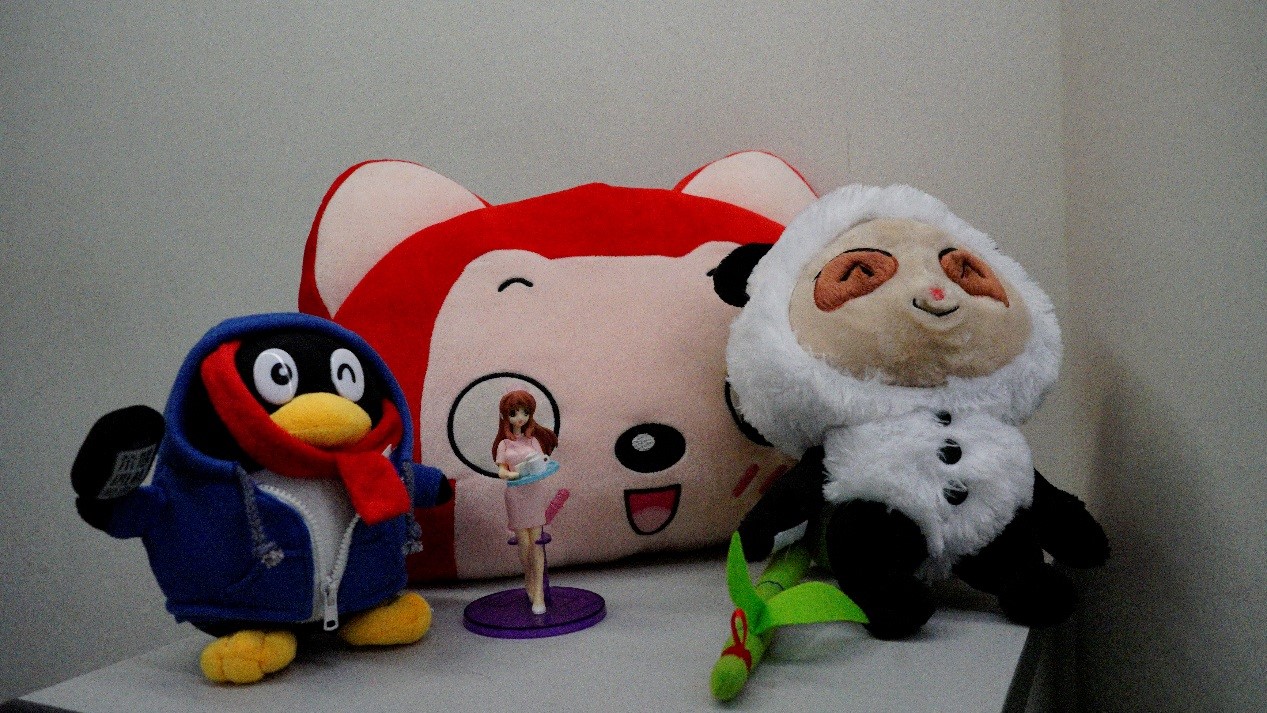}}
{\footnotesize (h) 6400,6.7,1/350}
\end{minipage}
\begin{minipage}[t]{0.32\textwidth}
\centering
\raisebox{-0.5cm}{\includegraphics[width=1\textwidth]{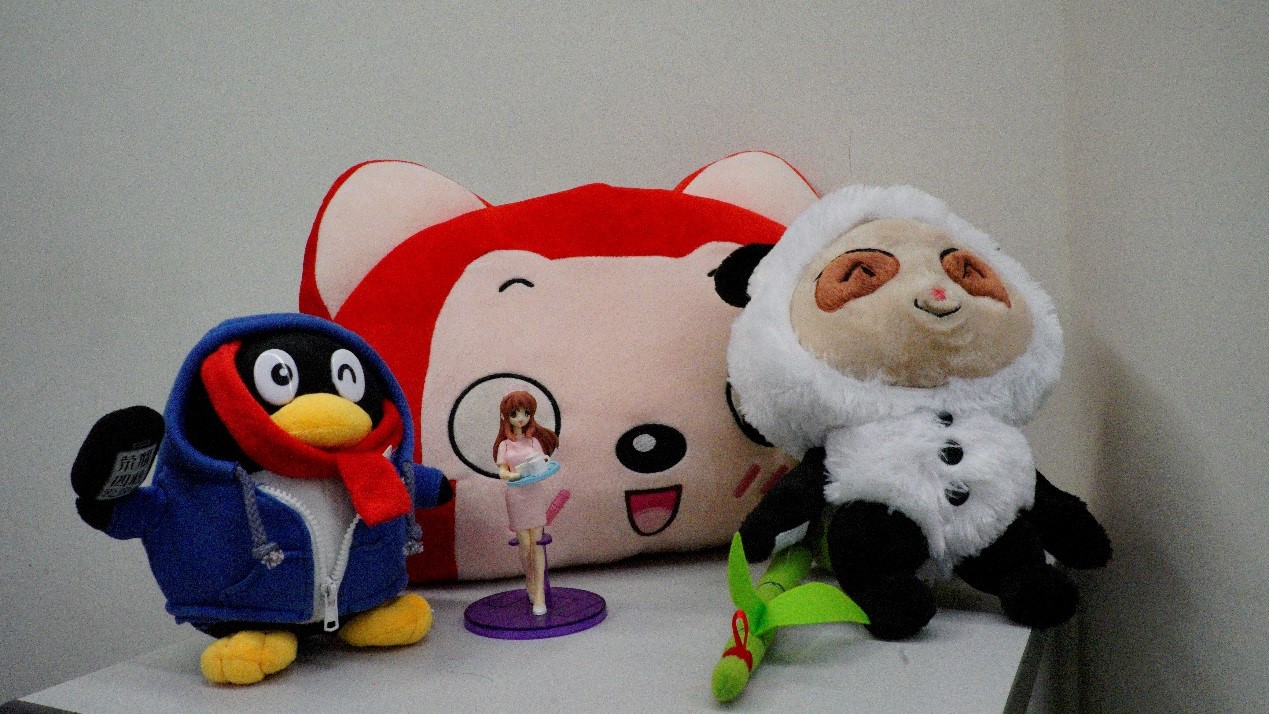}}
{\footnotesize (i) 6400,16,1/60}
\end{minipage}
}\vspace{-3mm}
    \caption{Captured images with the Sony A7 II camera under different (ISO, Shutter speed, Aperture) settings.}
    \label{fig6-1}
\end{figure*}

To analyze how ISO, shutter speed, and aperture influence the contents and illumination of the captured images, we perform some heuristic experiments with different camera settings. In Figure \ref{fig6-1}, we show some images captured with different camera settings. Comparing Figures \ref{fig6-1}(a) and \ref{fig6-1}(b) (or \ref{fig6-1}(g) and \ref{fig6-1}(i)), we can find that higher aperture results in darker illumination. Comparing Figures \ref{fig6-1}(b) and \ref{fig6-1}(c) (or \ref{fig6-1}(g) and \ref{fig6-1}(h)), we can find that slower shutter results in brighter illumination. Comparing Figrues \ref{fig6-1}(b), \ref{fig6-1}(d), \ref{fig6-1}(e), \ref{fig6-1}(f), and \ref{fig6-1}(g), we can find that the illumination becomes brighter when the ISO is higher. Besides, given fixed aperture and shutter, the images captured by the camera can avoid the over-exposure or under-exposure when the ISO is set between 400 and 3,200. When ISO=200, the captured images would have the problem of under-exposure, while when ISO=6,400, the captured images would have the problem of over-exposure. However, this can be alleviated by changing the aperture and shutter speed and finally we can obtain images with normal illuminations.

Given fixed shutter speed and aperture, enhancing the camera sensitivity will generate stronger noise than lowering the shutter speed. In the construction of our dataset, we only changed the ISO values while fixing shutter speed and aperture with suitable values to ensure that the images will not suffer over-exposure or under-exposure. It is commonly accepted that the noise in images will become stronger when the scene is under darker light conditions.

\subsection{The Dataset Construction Process}

\begin{figure*}[t!]
    \centering
\subfigure{
\begin{minipage}[t]{0.35\textwidth}
\centering
\raisebox{-0.5cm}{\includegraphics[width=1\textwidth]{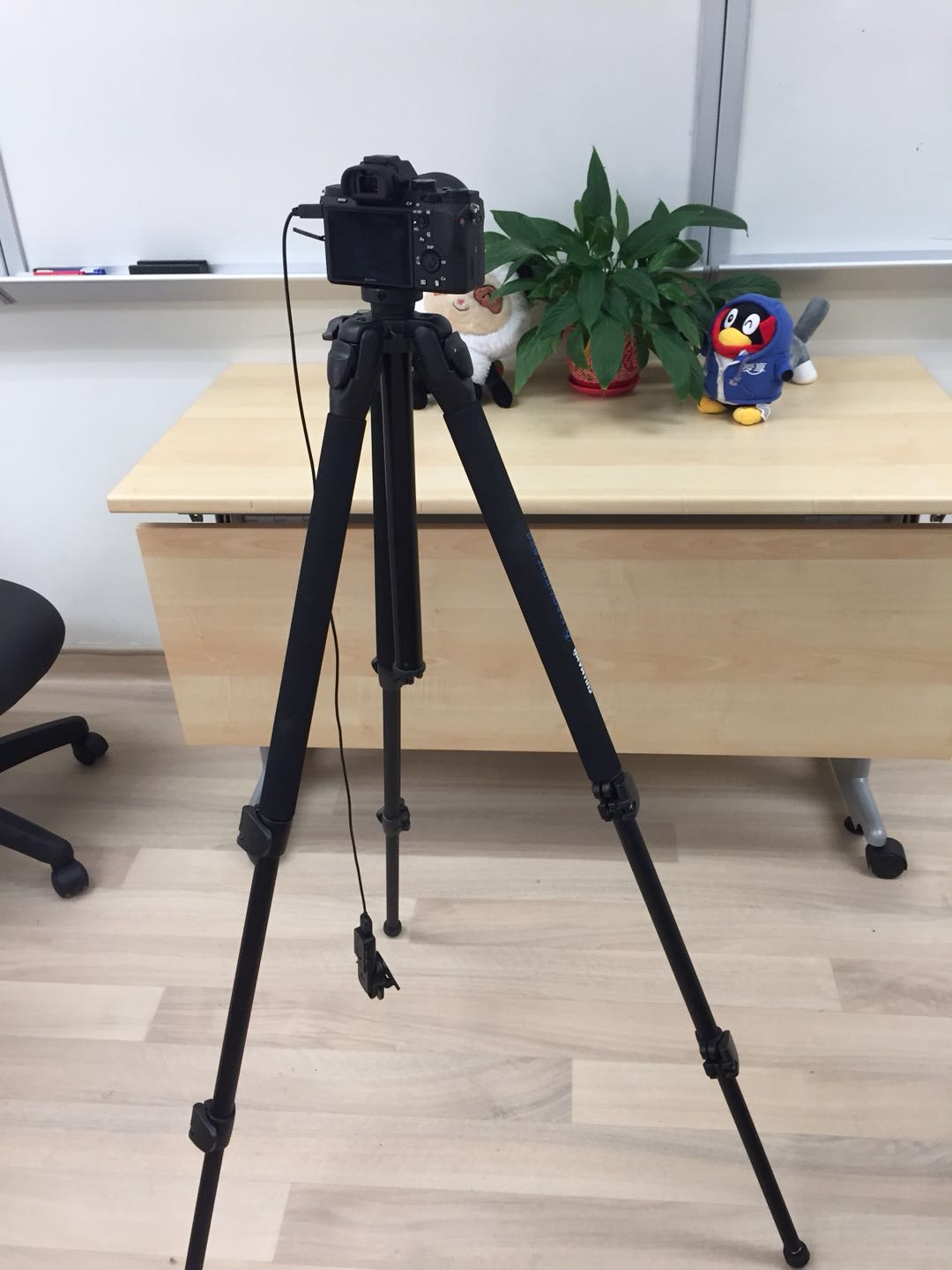}}
\end{minipage}
\begin{minipage}[t]{0.35\textwidth}
\centering
\raisebox{-0.5cm}{\includegraphics[width=1\textwidth]{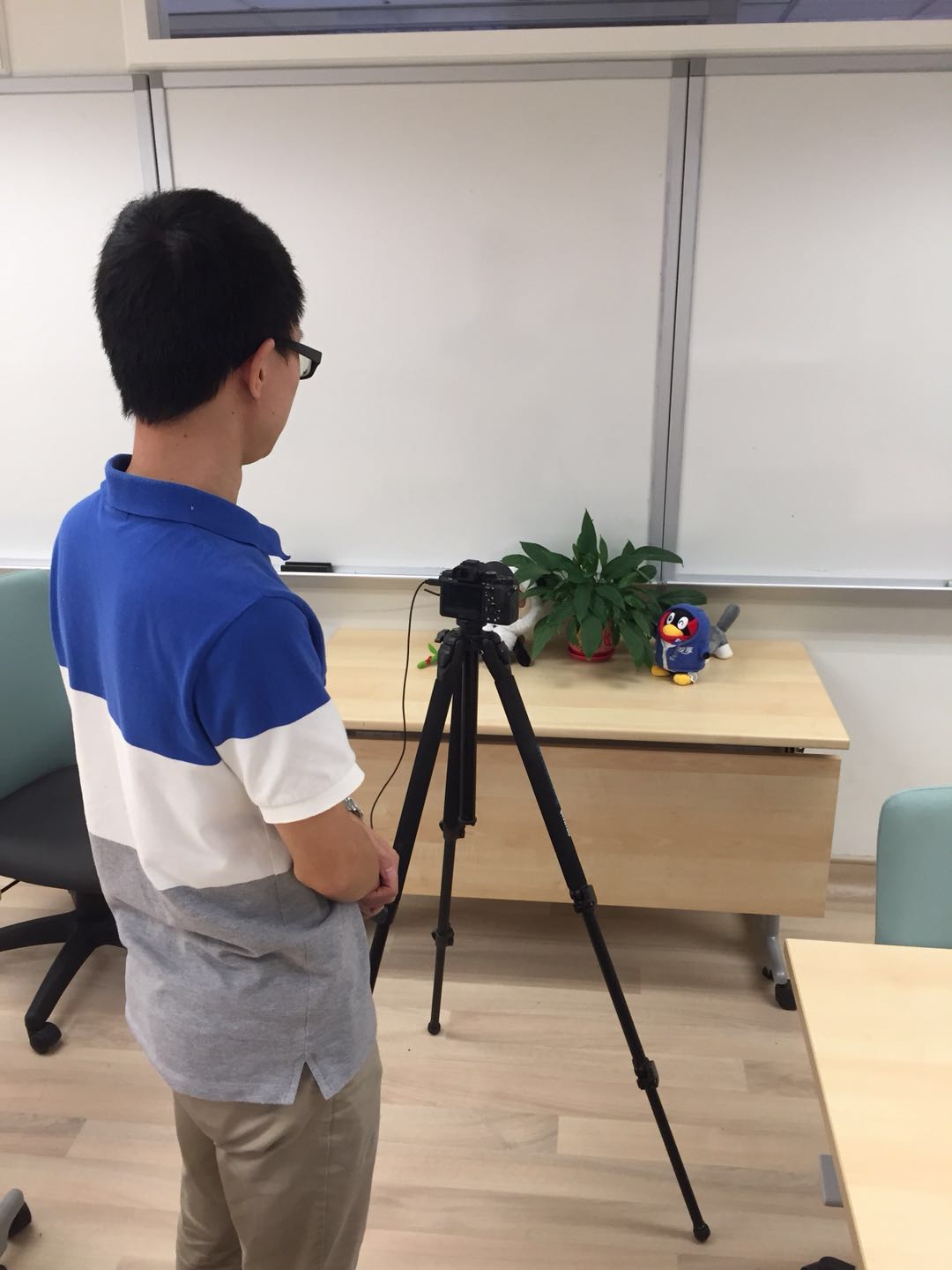}}
\end{minipage}
}\vspace{-3mm}
    \caption{The static scene is captured with a camera fixed by tripod. The data collection is automatically done with shutter release after the button is pressed by a person.}
    \label{fig6-2}
\end{figure*}

To alleviate the limitations of the previous datasets \cite{RENOIR2014,crosschannel2016,dnd2017}, we propose to construct a new dataset which could: 1) contain more camera brands; 2) contain more carefully designed camera settings; 3) capture more real-world scenes with realistic objects; 4) capture both the raw data and sRGB data for comparison analysis. The captured images are stored in raw data and JPEG format without compression. For each scene, we capture it for 500 times. Figure \ref{fig6-2} shows how we capture images of a static scens in indoor environment. The camera is fixed by a tripod. The data collection is automatically done with shutter release after the button is pressed by a person. Hence, the misalignment problem can be nearly avoided in the accquisition process of 500 images for one scene. We capture images with different camera settings. The cameras are set based on the following rules. First, the shutter speed should be faster than the blink of the fluorescent lights, otherwise the flickering of the light will make the global luminances of the captured images very different. Second, we set the shutter speed, the aperture, and the ISO value to ensure that the scenes are in a naturally lighting condition. Besides, since the digital single-lens reflex cameras (DSLRs) use mechanical shutter, the shutter speed of each shot is a little different. This small difference results in slightly different brightness of different shots. However, we ignore this small difference in our dataset, as that in \cite{crosschannel2016}.

\textbf{More Camera Brands}: In our dataset, we use 5 different cameras of three camera brands, including Canon (Mark 5D, 80D, 600D), Nikon (D800), and Sony (A7 II), to capture real-world noisy images. According to a recent survey \cite{commoncamera}, the three camera brands occupy 48 of 50 most commonly used camera-lens combinations. Hence, our dataset is more comprehensive than the previous datasets on camera brands.

% The other two brands are from the camera brands of Fujifilm and Olympus.

\begin{figure*}%[t!]
    \centering
\subfigure{
\begin{minipage}[t]{0.3\textwidth}
\centering
\raisebox{-0.5cm}{\includegraphics[width=1\textwidth]{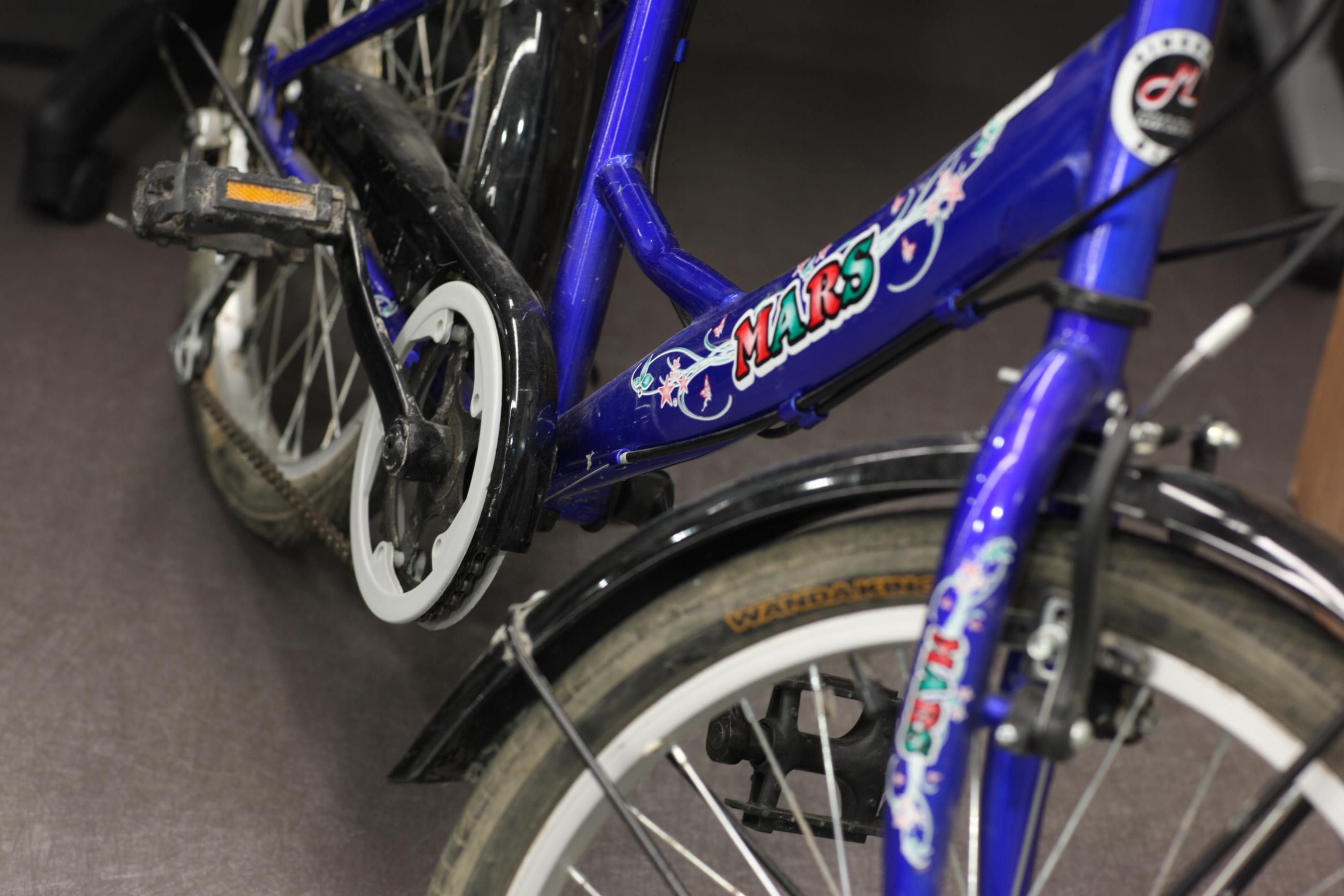}}
\end{minipage}
\begin{minipage}[t]{0.3\textwidth}
\centering
\raisebox{-0.5cm}{\includegraphics[width=1\textwidth]{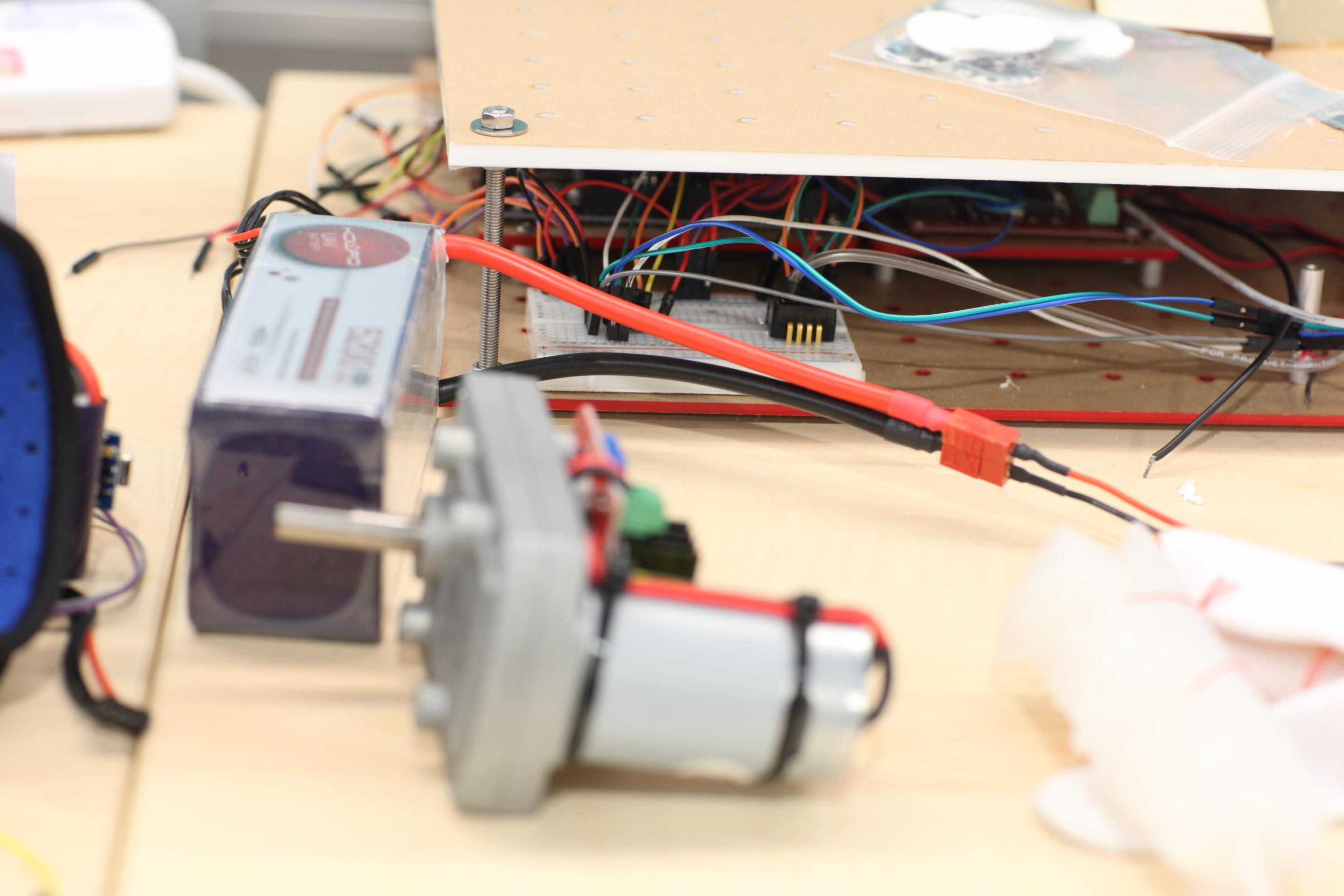}}
\end{minipage}
\begin{minipage}[t]{0.3\textwidth}
\centering
\raisebox{-0.5cm}{\includegraphics[width=1\textwidth]{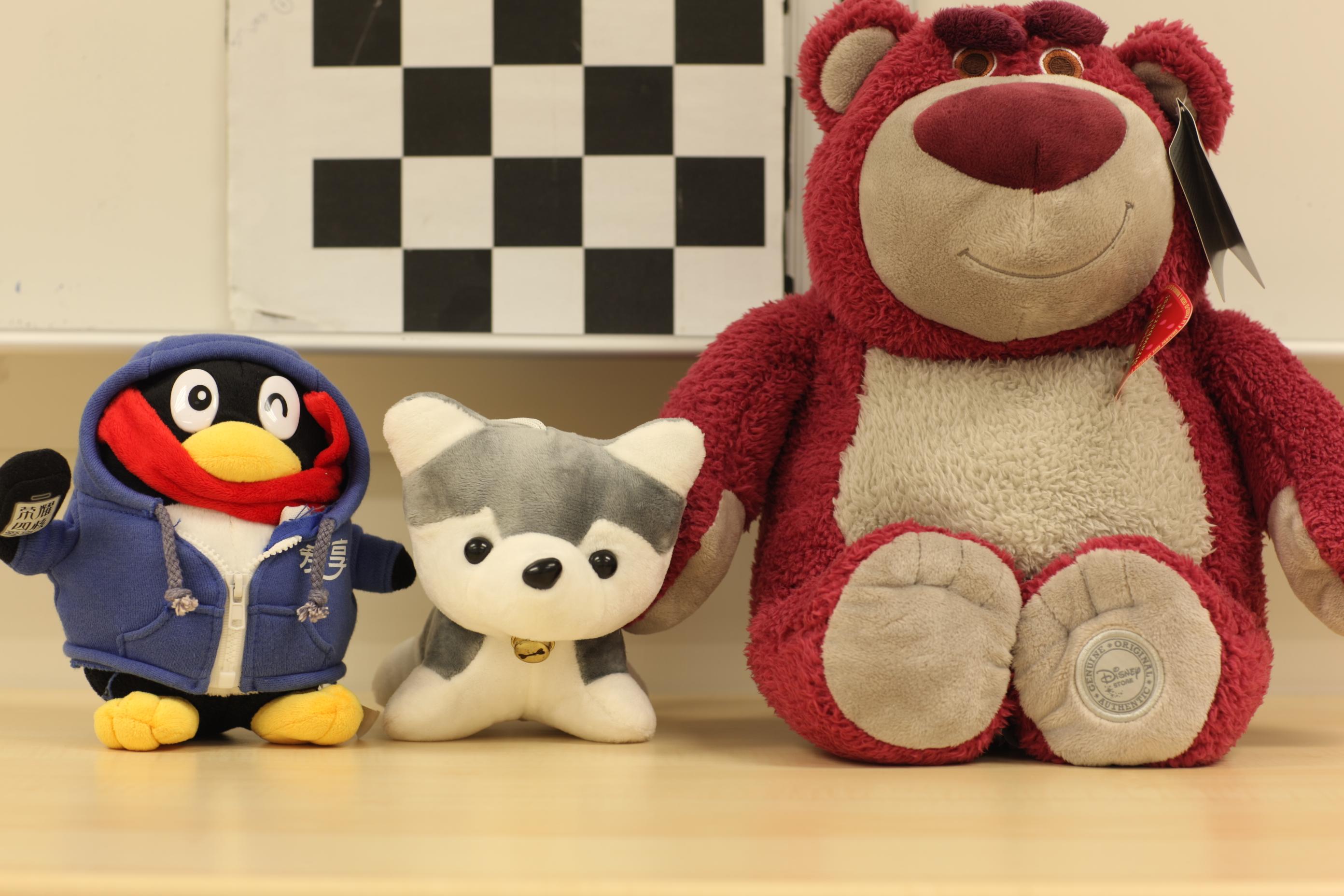}}
\end{minipage}
}\vspace{-3mm}
\subfigure{
\begin{minipage}[t]{0.3\textwidth}
\centering
\raisebox{-0.5cm}{\includegraphics[width=1\textwidth]{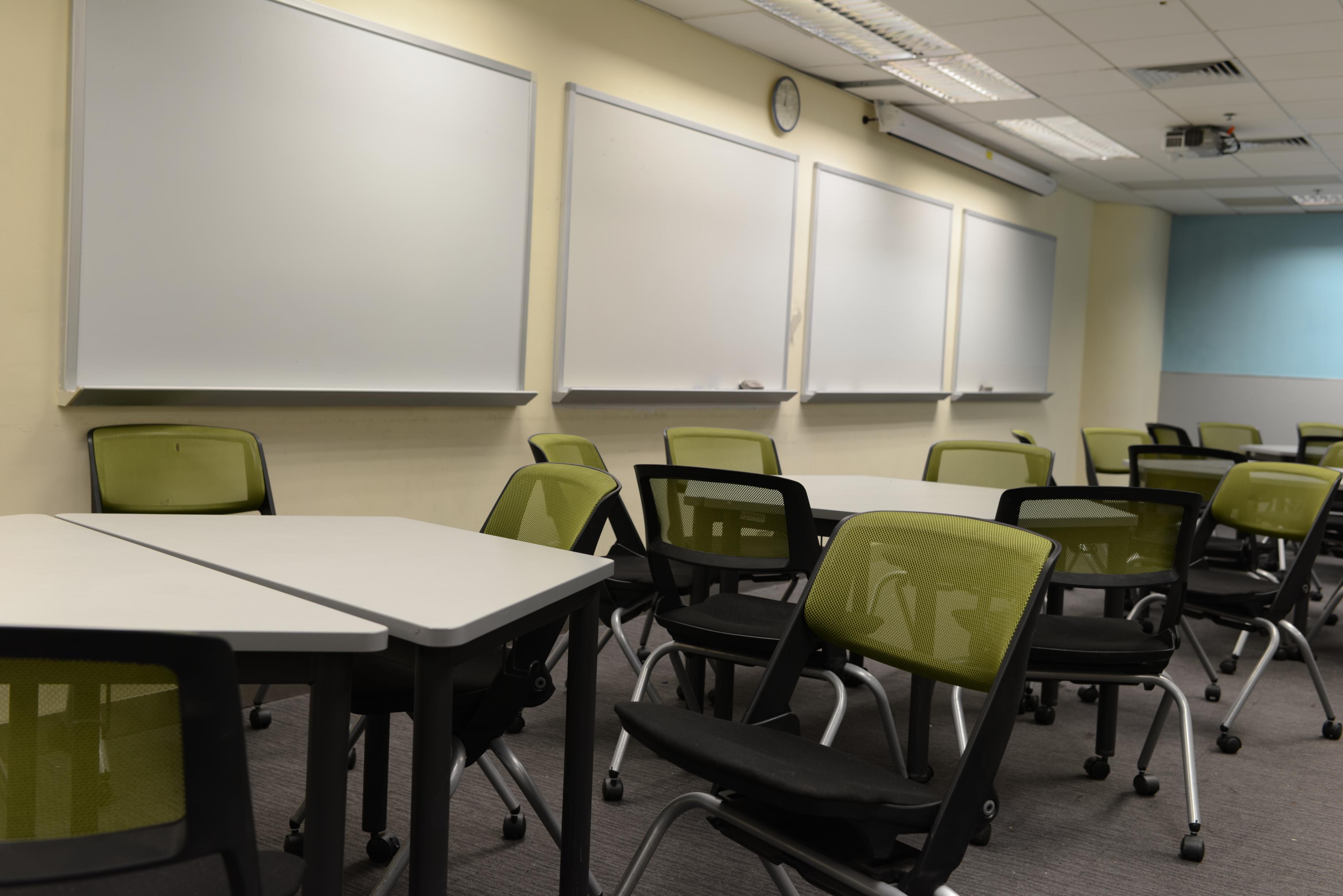}}
\end{minipage}
\begin{minipage}[t]{0.3\textwidth}
\centering
\raisebox{-0.5cm}{\includegraphics[width=1\textwidth]{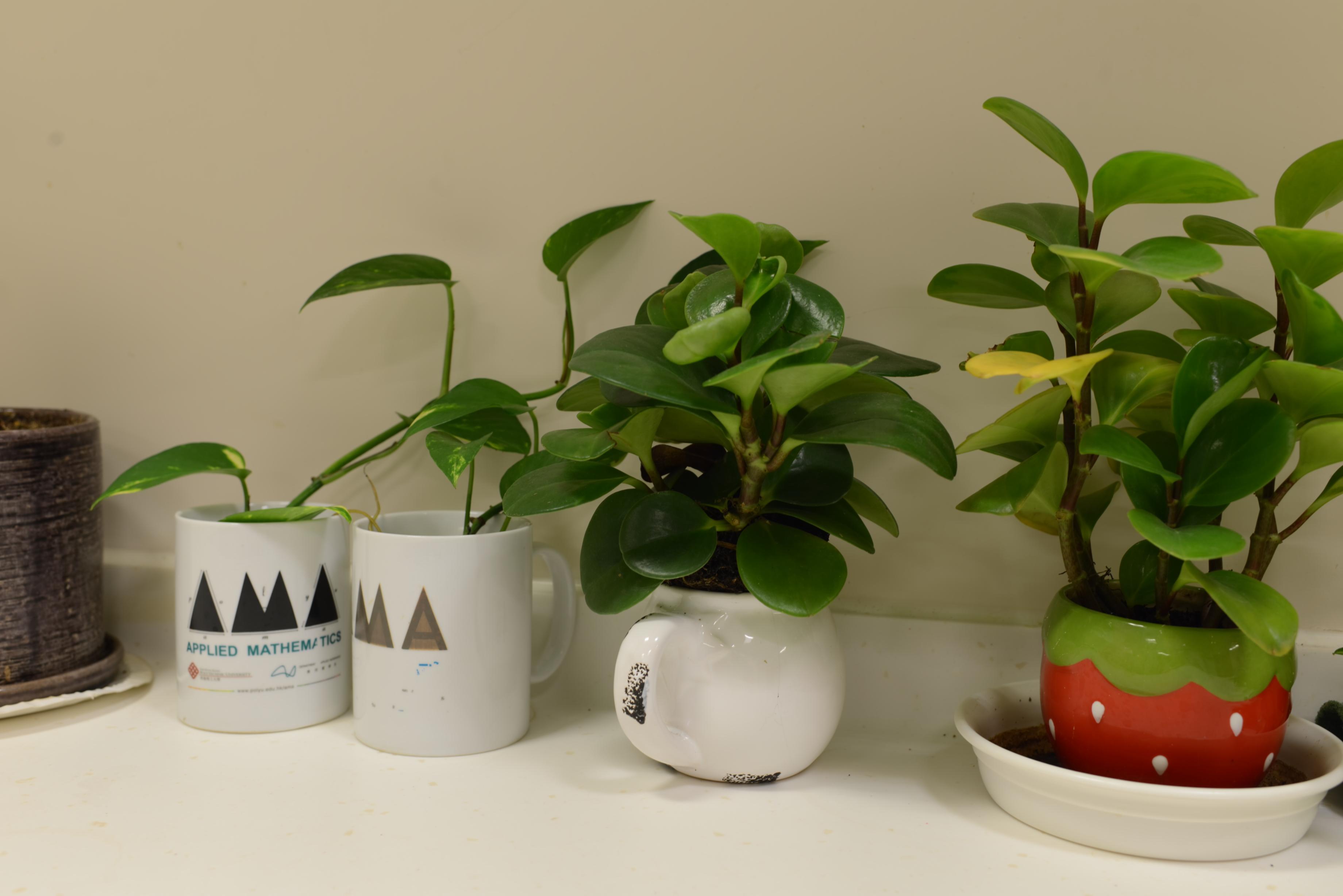}}
\end{minipage}
\begin{minipage}[t]{0.3\textwidth}
\centering
\raisebox{-0.5cm}{\includegraphics[width=1\textwidth]{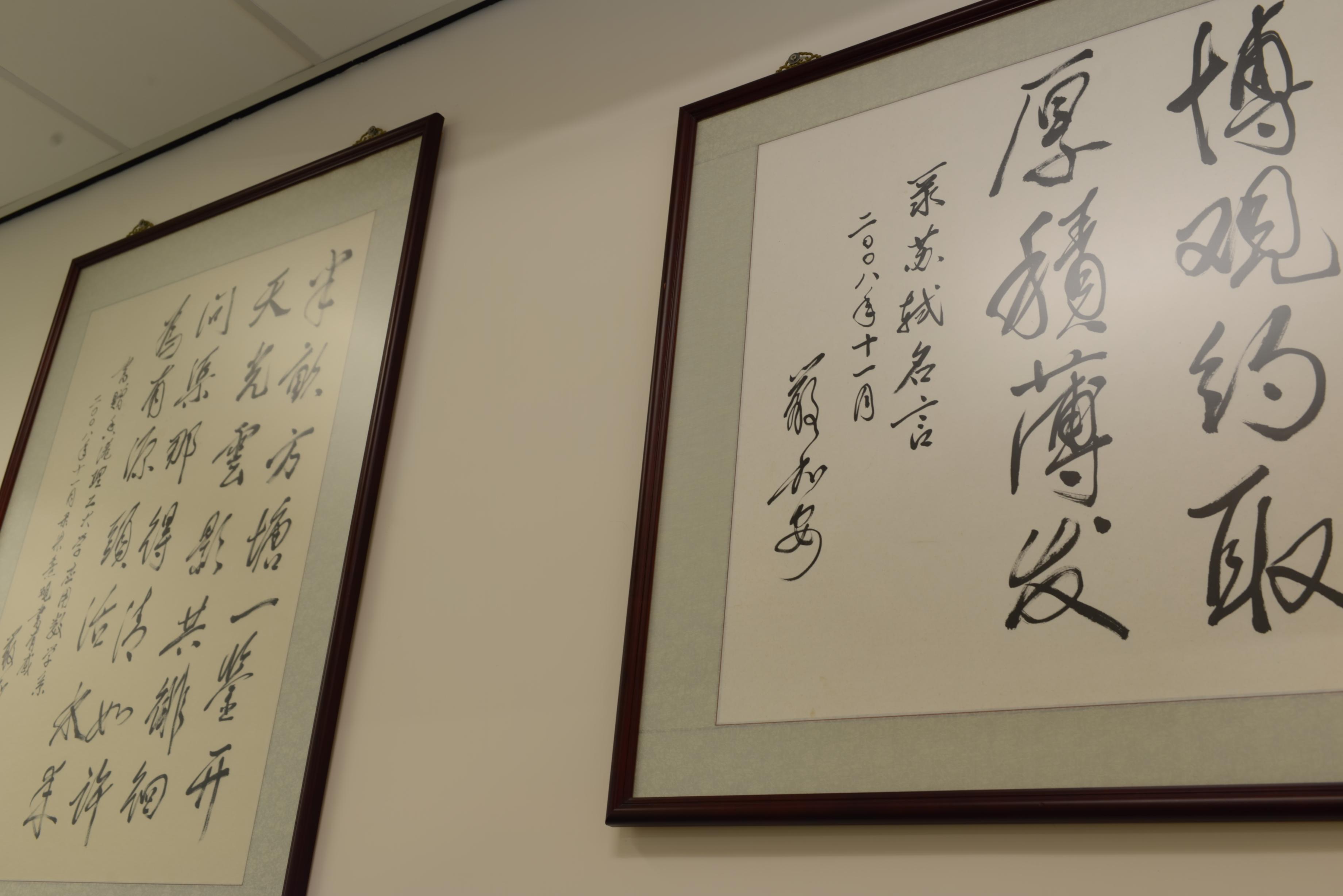}}
\end{minipage}
}\vspace{-3mm}
\subfigure{
\begin{minipage}[t]{0.3\textwidth}
\centering
\raisebox{-0.5cm}{\includegraphics[width=1\textwidth]{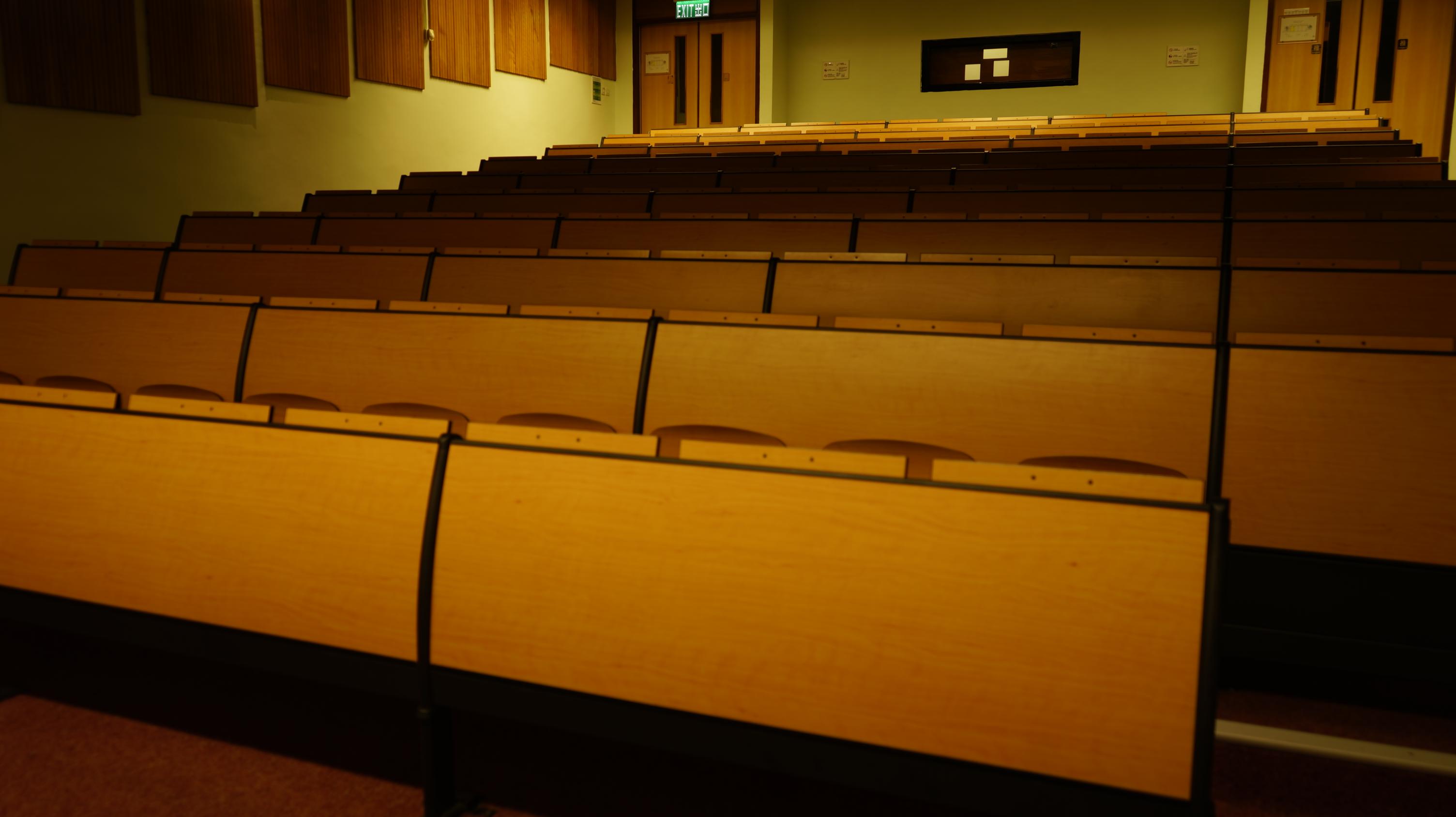}}
\end{minipage}
\begin{minipage}[t]{0.3\textwidth}
\centering
\raisebox{-0.5cm}{\includegraphics[width=1\textwidth]{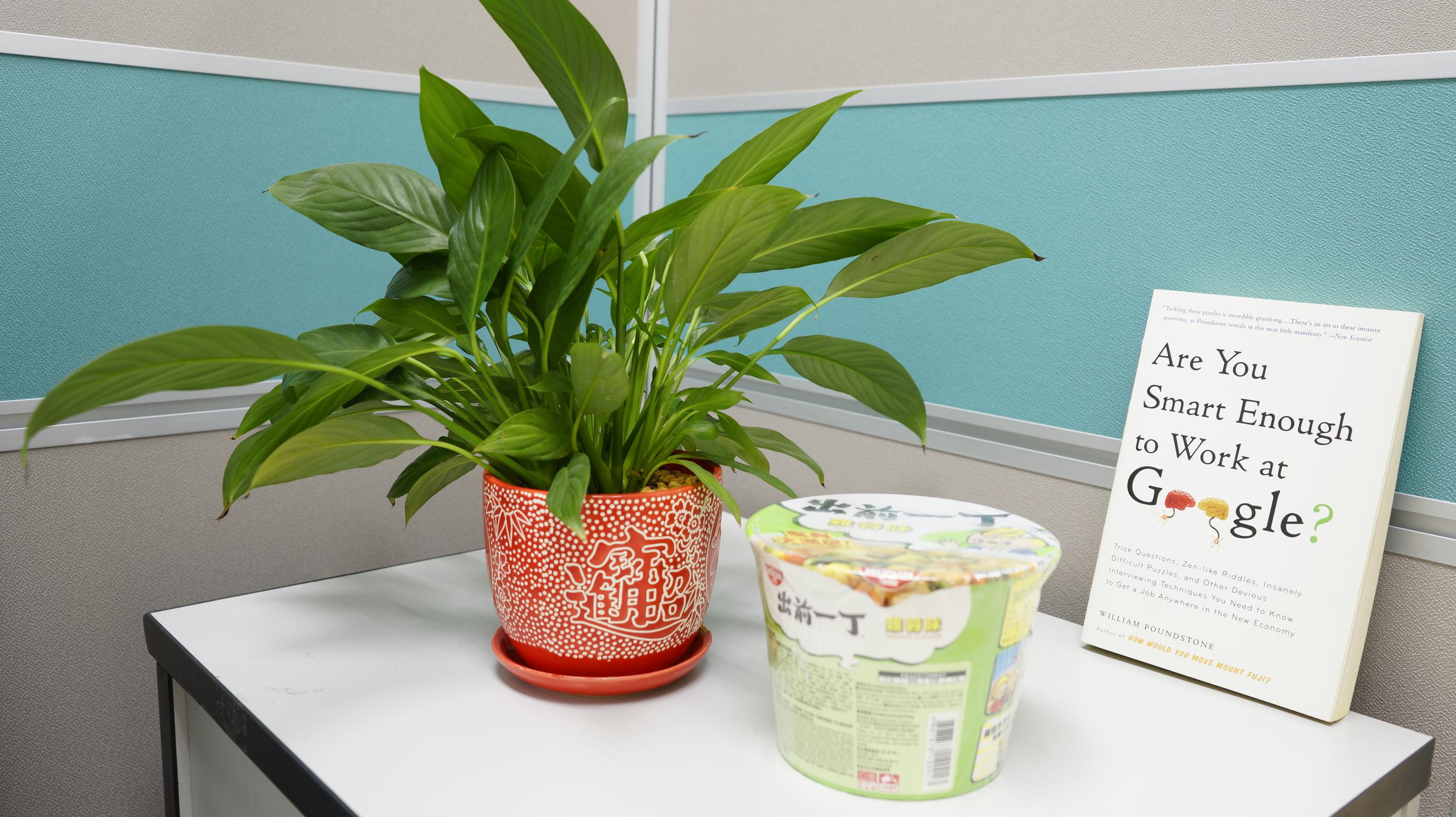}}
\end{minipage}
\begin{minipage}[t]{0.3\textwidth}
\centering
\raisebox{-0.5cm}{\includegraphics[width=1\textwidth]{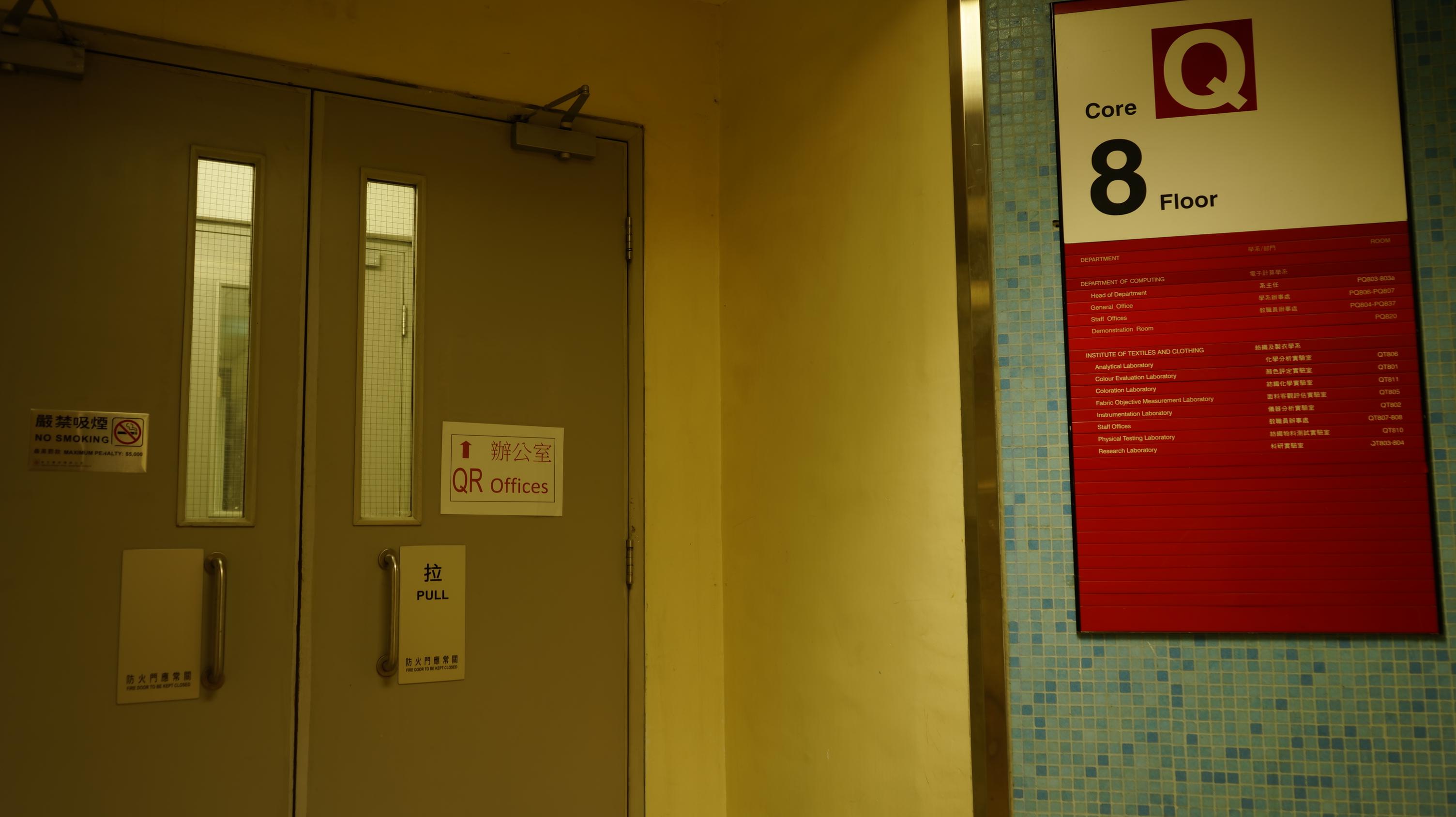}}
\end{minipage}
}\vspace{-3mm}
\subfigure{
\begin{minipage}[t]{0.3\textwidth}
\centering
\raisebox{-0.5cm}{\includegraphics[width=1\textwidth]{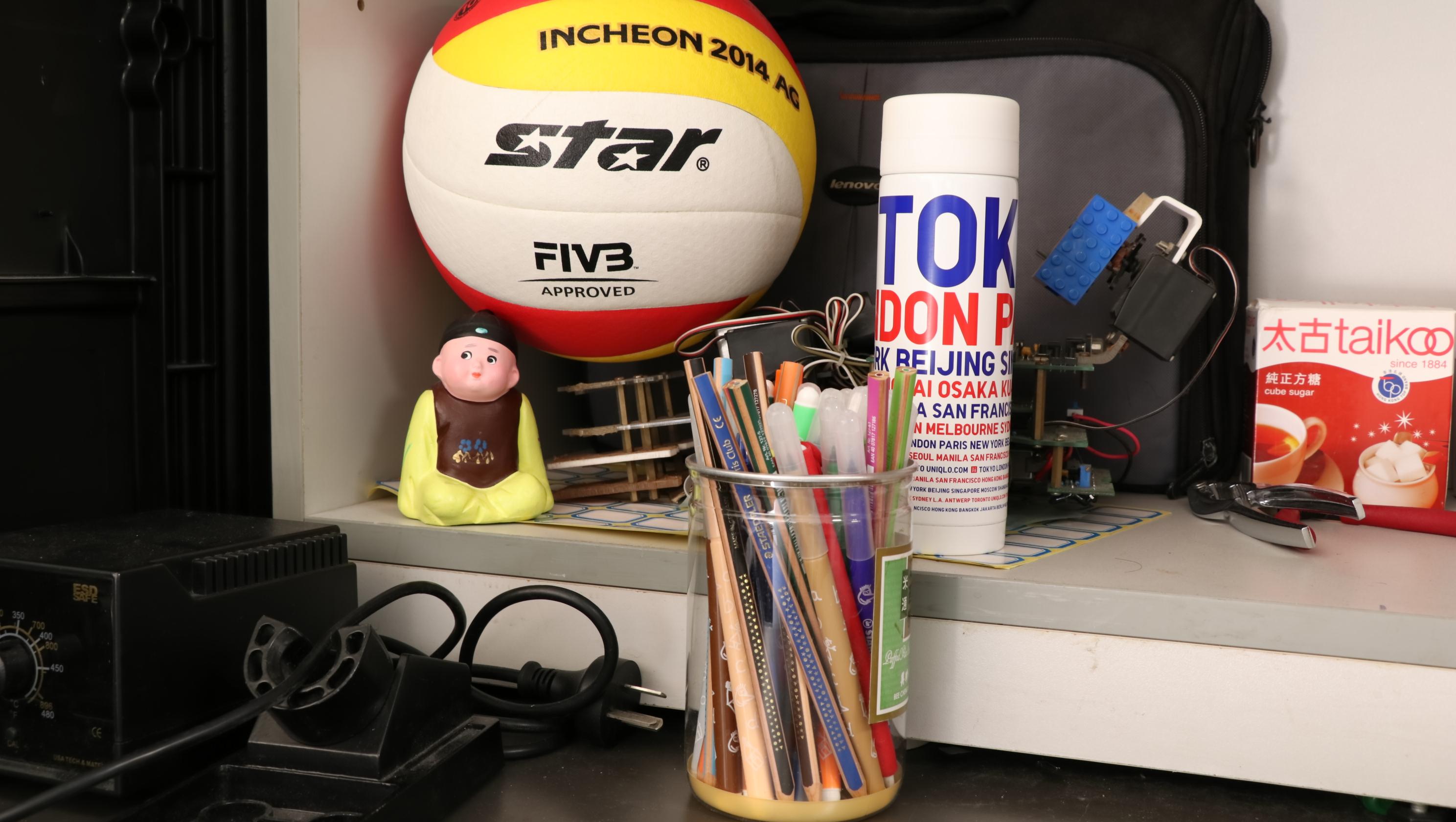}}
\end{minipage}
}\vspace{-3mm}
\subfigure{
\begin{minipage}[t]{0.3\textwidth}
\centering
\raisebox{-0.5cm}{\includegraphics[width=1\textwidth]{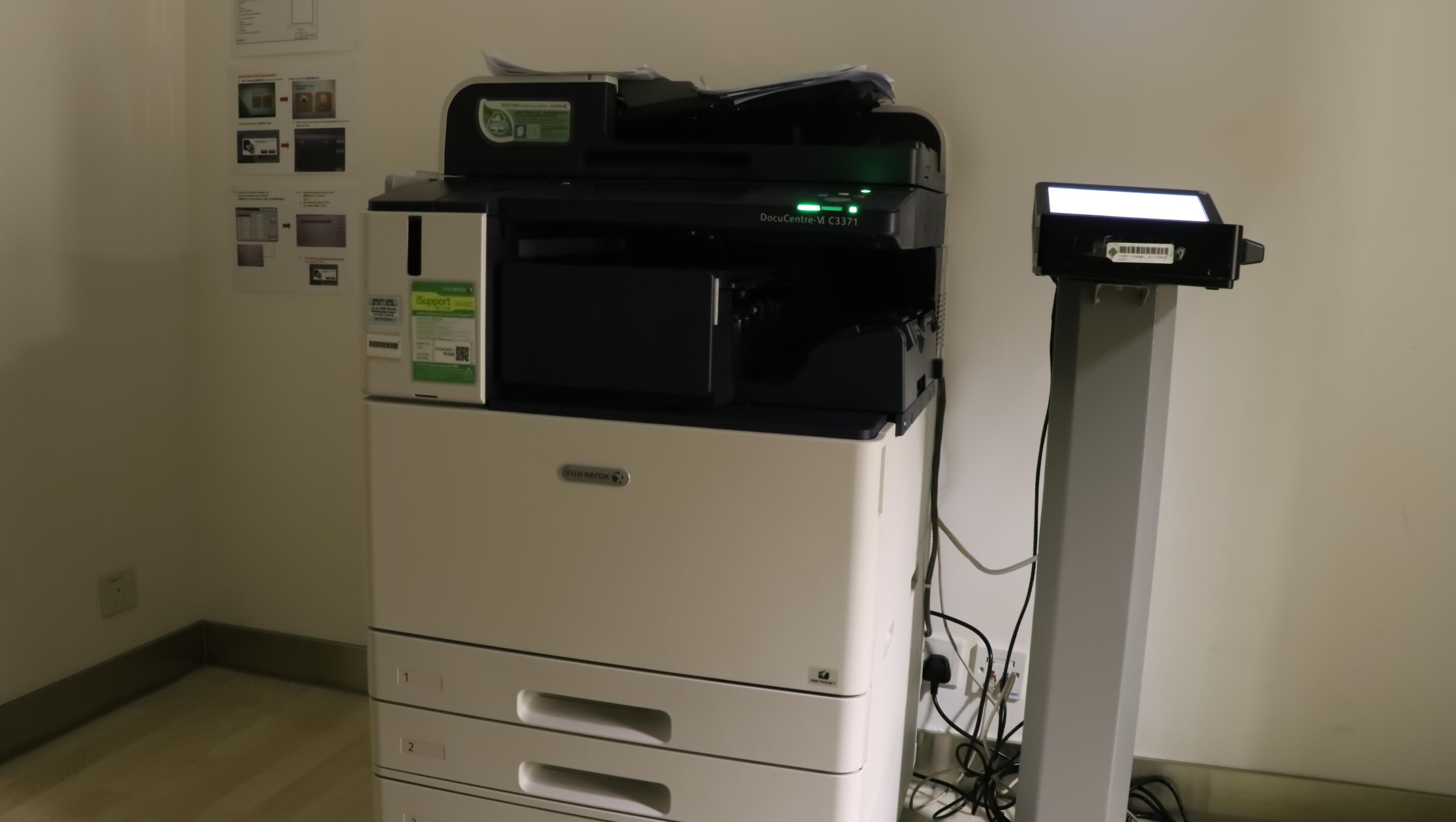}}
\end{minipage}
\begin{minipage}[t]{0.3\textwidth}
\centering
\raisebox{-0.5cm}{\includegraphics[width=1\textwidth]{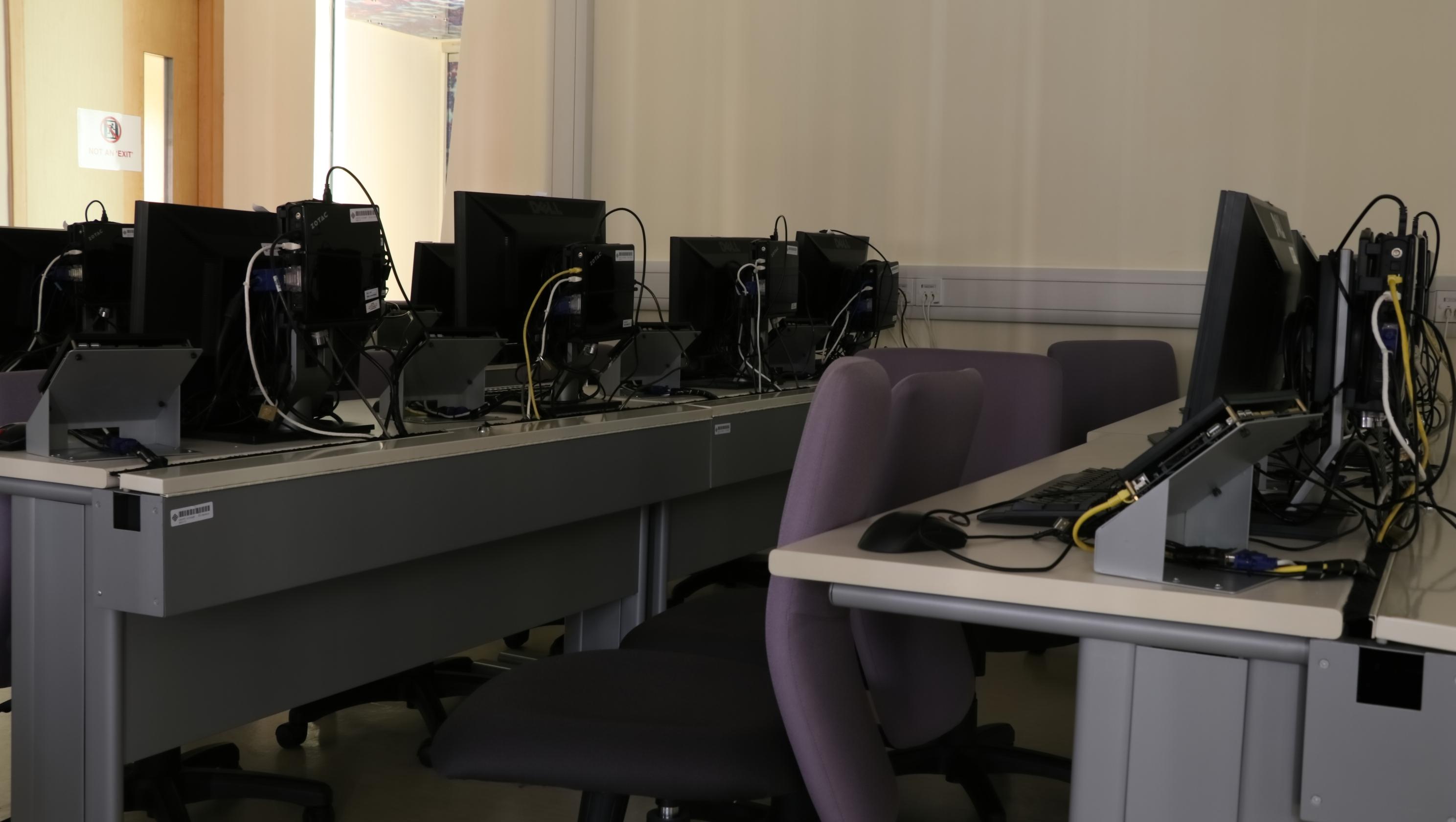}}
\end{minipage}
}
\caption{Some sample images in our newly constructed dataset.}
\label{fig6-3}
\end{figure*}

\textbf{More Camera Settings}: In our new dataset, each scene is captured with 6 different ISO settings, e.g., 800, 1,600, 3,200, 6,400, 12,800, and 25,600. For each ISO setting, we carefully adjust the shutter speed and aperture, and choose other suitable camera settings to make the captured scene neither under-exposed nor over-exposed. With the increase of ISO, the luminance of the captured images will also increase, and we tune the shutter speed and apeture accordingly to capture an image with normal luminance. For example, to make the images captured with ISO=25,600 be normally exposed, we set the shutter speed to 1/320 second and aperture to F10.0.

\textbf{More Captured Scenes}: We capture the images with indoor normal lighting condition, dark lighting condition, and outdoor normal lighting condition. The scenes we captured are also versaltile (including the buildings, classrooms, caffe rooms, and outdoor scenes, etc.). The objects in the scenes include books, pens, bottles, boxes, and joys, etc. In summary, we capture totally 40 different scenes by using 5 different cameras in different camera settings, including 12 scenes captured by Canon 5D Mark II, 5 scenes captured by Canon 80D, 3 scenes captured by Canon 600D, 13 scenes captured by Nikon D800, 7 scenes captured by Sony A7 II. Since the images are of large size ($3000\times3000$), we crop some regions from these images and obtain 100 regions of size $512\times512$.

\textbf{Removing Outlier Images}: The outlier images are those images which have misalignment or different illuminance from the base image (we usually choose the first image of the 500 shots as the base image). In the dataset of \cite{crosschannel2016}, the authors did not remove the images with misalignment or different luminances. In the DND dataset \cite{dnd2017}, the authors corrected the misalignment of each image. However, this operation largely depends on the misalignment detection method and the correction method, which may make the corrected images less natural. Besides, the DND dataset \cite{dnd2017} takes the image captured with low ISO as ``ground truth'', and linearly transfer the noisy image captured with high ISO to the scale of the ``ground truth'' image. This step, in our opinion, is problematic since the image pixels are not linearly dependent on the ISO values. In our dataset, we browse the captured images and remove the outlier images with clear misalignment or different luminances. For each scene, three volunteers are invited to do the screening successively, and the remaining images are used to compute the ``ground truth'' image.

\textbf{Generating ``Ground Truth'' Image}: The ``ground truth'' images of the RENOIR dataset \cite{RENOIR2014} are generated when the camera is set with ISO=100, while the other settings are fixed the same as those for the noisy images. The ``ground truth'' images of the dataset \cite{crosschannel2016} are generated by averaging the static images captured on the same scene under the same camera setting. The ``ground truth'' images of the DND dataset \cite{dnd2017} is generated mainly by using low ISO values (e.g., ISO=100), and other post-processing steps include linear intensity changes, spatial misalignment, and low-frequency residual correciton, etc. In our dataset, we employ the same strategy as that method of \cite{crosschannel2016} due to its simplisity. We capture images of the same static scenes for many ($500\sim1000$) times and average the captured images to obtain the ``ground truth'' image. 

We first remove the images with misalignment by careful subjective evaluation. The images with several pixels displacement will be deleted. After this stage, we will remove the images with inconsistant luminance. The luminance is affected by two factors. One is the lighting conditions of the environment. The other is that the camera will automatically make up the illumination when the scene is in a relatively dark lighting condition. Since we shot the scene for many times, some shots may have different illumination, though captured under the same lighting condition. To remove the images with outlier luminance, we first sample 10,000 pixels uniformly (the pixels are on the 100 equidistantly sampled rows and 100 equidistantly sampled columns) from each image, and then compute the mean luminance of the 10,000 pixels. Each of the captured images will have one value representing its mean luminance. We sort these values in a descending order. The images with the lowest or highest mean luminances will be referred as outlier images. We remove these images until the lowest and highest mean luminances are close enough to the ``center'' of the mean luminances. Here, ``center'' means the median of the sorted mean luminances. In this way, the images which are much darker or much brighter than the ``center'' image with the ``center'' luminance will be deleted, and the remaining images are very close to each other in luminance. The remaining images will be averaged to obtain the mean image, which will be used as the ``ground truth'' image of each scene.

\subsection{Summary of the Dataset}

In our constructed dataset, we captured images from 40 different scenes with different contents and objects. Figure \ref{fig6-3} shows some samples of the real-world noisy images in our dataset. The images cover from different types of indoor scenes and versaitile objects, etc.

\begin{figure*}%[ht!]
\centering
\subfigure{
\begin{minipage}[t]{0.24\textwidth}
\centering
\raisebox{-0.5cm}{\includegraphics[width=1\textwidth]{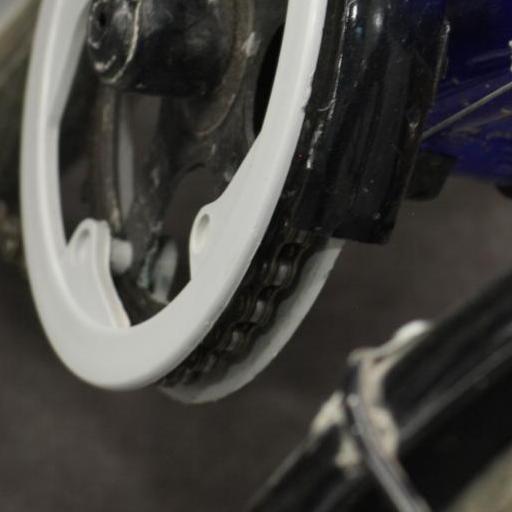}}
\end{minipage}
\begin{minipage}[t]{0.24\textwidth}
\centering
\raisebox{-0.5cm}{\includegraphics[width=1\textwidth]{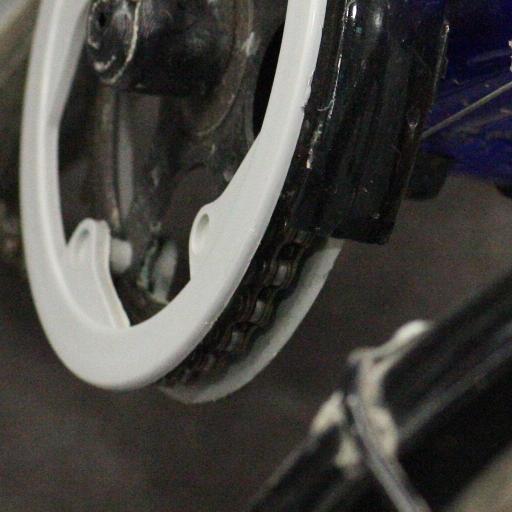}}
\end{minipage}
\begin{minipage}[t]{0.24\textwidth}
\centering
\raisebox{-0.5cm}{\includegraphics[width=1\textwidth]{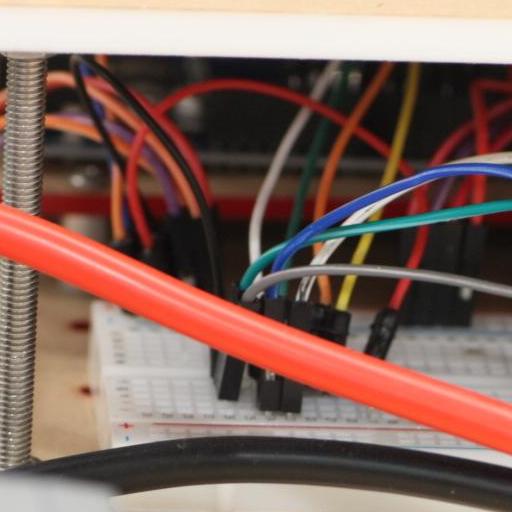}}
\end{minipage}
\begin{minipage}[t]{0.24\textwidth}
\centering
\raisebox{-0.5cm}{\includegraphics[width=1\textwidth]{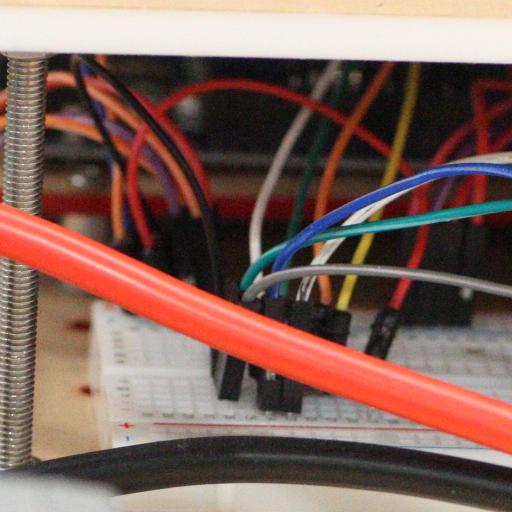}}
\end{minipage}
}
\subfigure{
\begin{minipage}[t]{0.24\textwidth}
\centering
\raisebox{-0.5cm}{\includegraphics[width=1\textwidth]{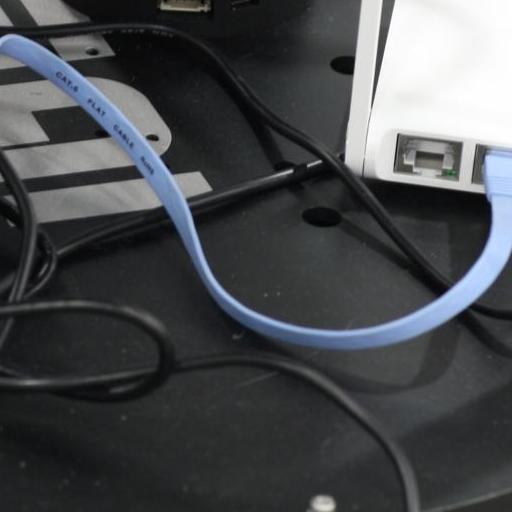}}
\end{minipage}
\begin{minipage}[t]{0.24\textwidth}
\centering
\raisebox{-0.5cm}{\includegraphics[width=1\textwidth]{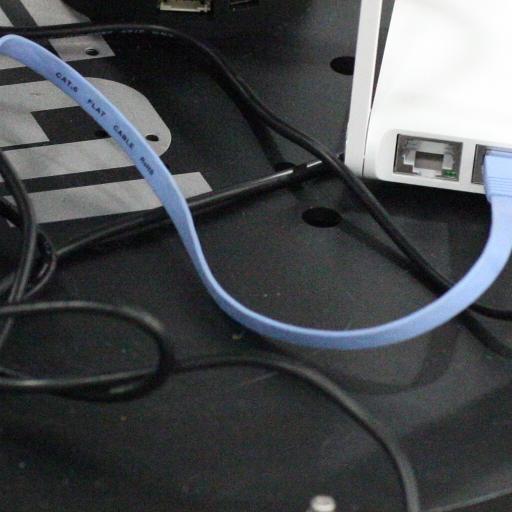}}
\end{minipage}
\begin{minipage}[t]{0.24\textwidth}
\centering
\raisebox{-0.5cm}{\includegraphics[width=1\textwidth]{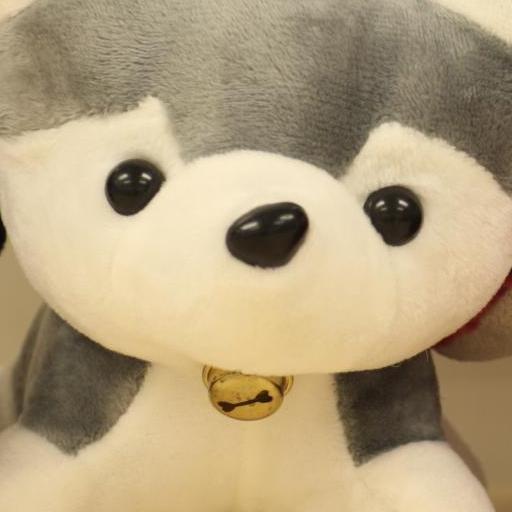}}
\end{minipage}
\begin{minipage}[t]{0.24\textwidth}
\centering
\raisebox{-0.5cm}{\includegraphics[width=1\textwidth]{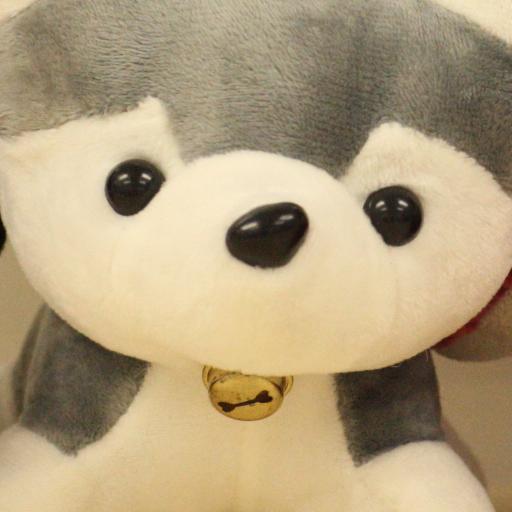}}
\end{minipage}
}
\subfigure{
\begin{minipage}[t]{0.24\textwidth}
\centering
\raisebox{-0.5cm}{\includegraphics[width=1\textwidth]{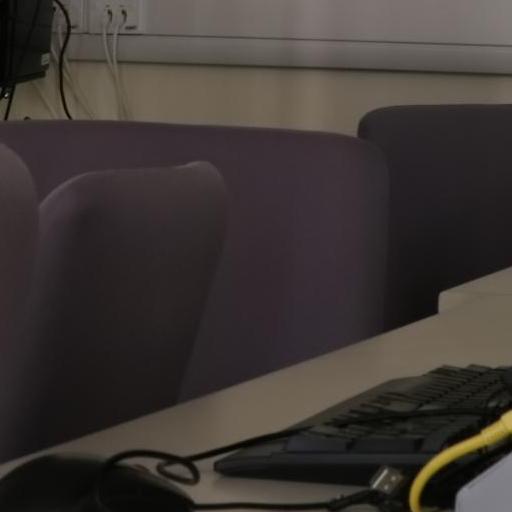}}
\end{minipage}
\begin{minipage}[t]{0.24\textwidth}
\centering
\raisebox{-0.5cm}{\includegraphics[width=1\textwidth]{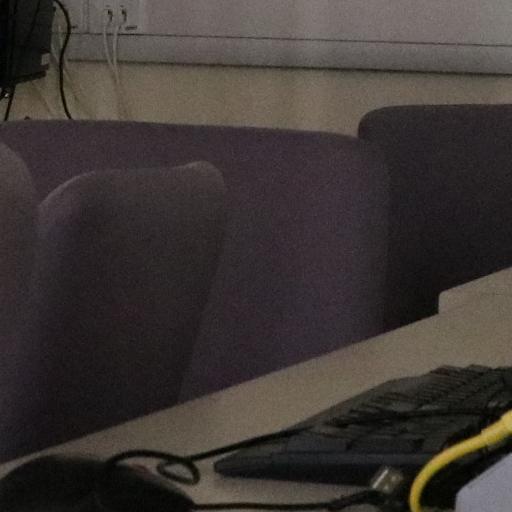}}
\end{minipage}
\begin{minipage}[t]{0.24\textwidth}
\centering
\raisebox{-0.5cm}{\includegraphics[width=1\textwidth]{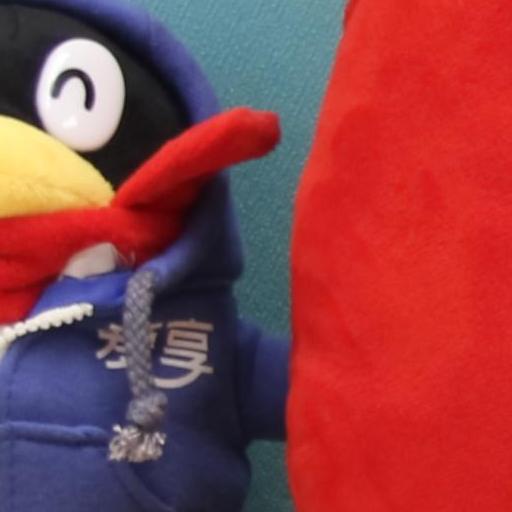}}
\end{minipage}
\begin{minipage}[t]{0.24\textwidth}
\centering
\raisebox{-0.5cm}{\includegraphics[width=1\textwidth]{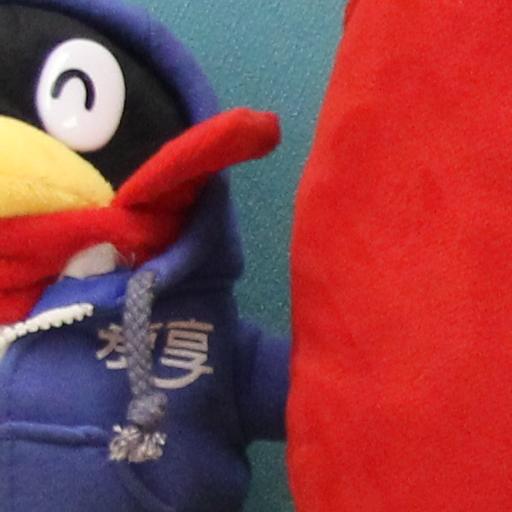}}
\end{minipage}
}
\subfigure{
\begin{minipage}[t]{0.24\textwidth}
\centering
\raisebox{-0.5cm}{\includegraphics[width=1\textwidth]{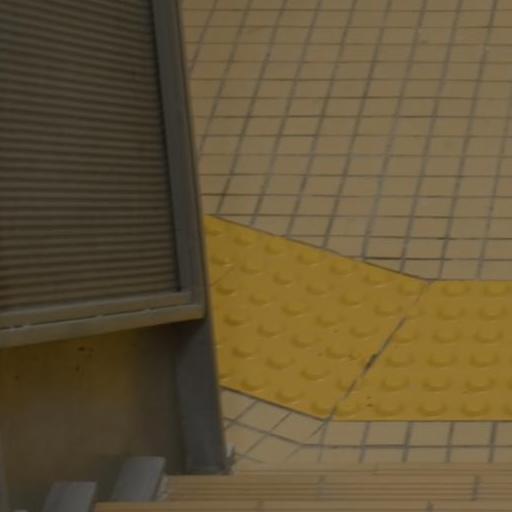}}
\end{minipage}
\begin{minipage}[t]{0.24\textwidth}
\centering
\raisebox{-0.5cm}{\includegraphics[width=1\textwidth]{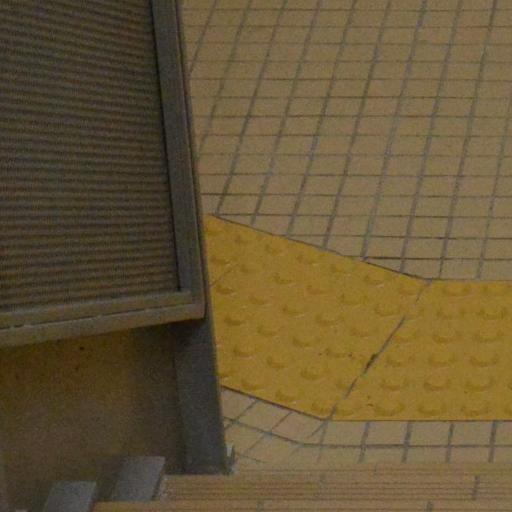}}
\end{minipage}
\begin{minipage}[t]{0.24\textwidth}
\centering
\raisebox{-0.5cm}{\includegraphics[width=1\textwidth]{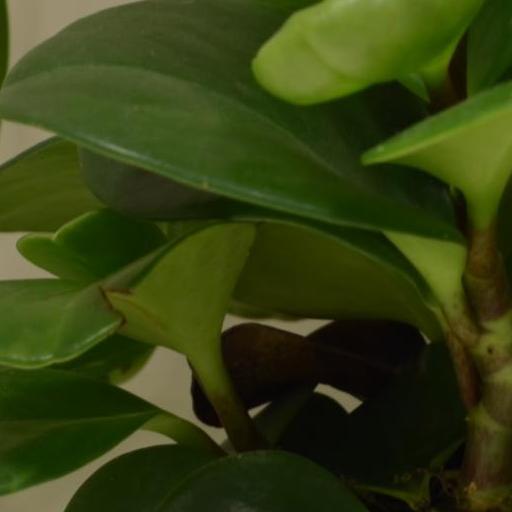}}
\end{minipage}
\begin{minipage}[t]{0.24\textwidth}
\centering
\raisebox{-0.5cm}{\includegraphics[width=1\textwidth]{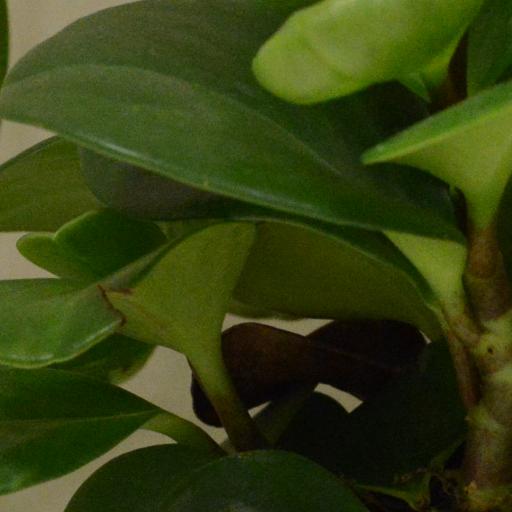}}
\end{minipage}
}
\subfigure{
\begin{minipage}[t]{0.24\textwidth}
\centering
\raisebox{-0.5cm}{\includegraphics[width=1\textwidth]{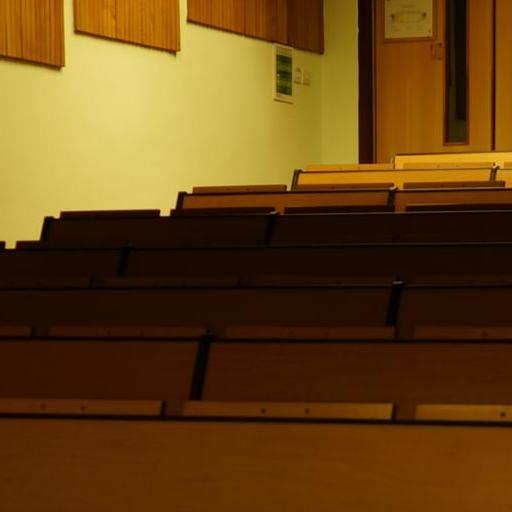}}
\end{minipage}
\begin{minipage}[t]{0.24\textwidth}
\centering
\raisebox{-0.5cm}{\includegraphics[width=1\textwidth]{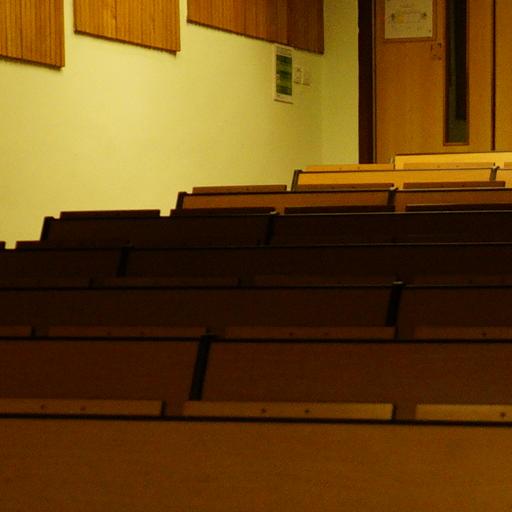}}
\end{minipage}
\begin{minipage}[t]{0.24\textwidth}
\centering
\raisebox{-0.5cm}{\includegraphics[width=1\textwidth]{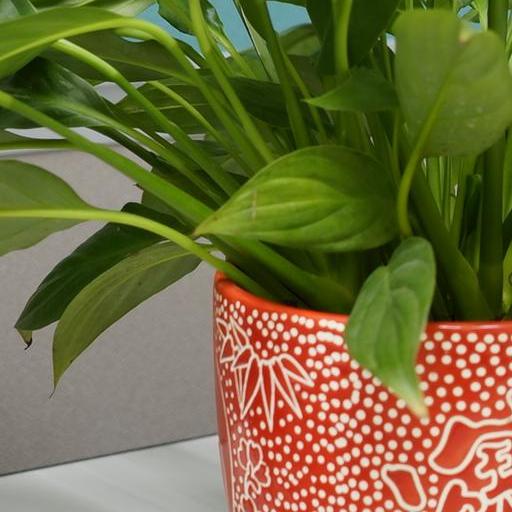}}
\end{minipage}
\begin{minipage}[t]{0.24\textwidth}
\centering
\raisebox{-0.5cm}{\includegraphics[width=1\textwidth]{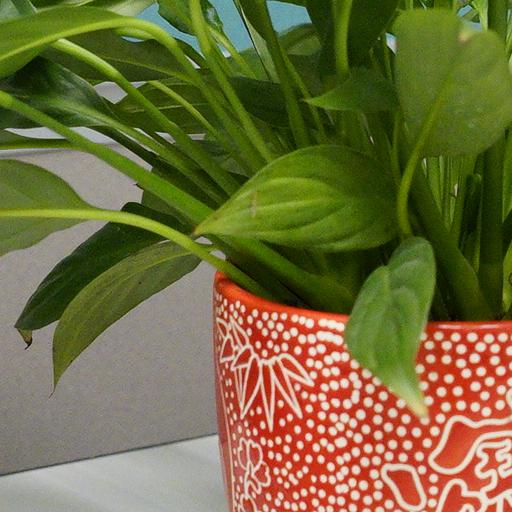}}
\end{minipage}
}
    \caption{Some cropped regions of the ``ground truth'' images (left) and their corresponding noisy images (right) in our constructed dataset.}
    \label{fig6-4}
\end{figure*}

Since the images we captured are very large in size, we crop 100 regions of size $512\times512$ from the 40 scenes to evaluate the existing image denoising methods. Some examples of the cropped regions and their corresponding ``ground truth'' images are listed in Figure \ref{fig6-4}. Besides, one can see that the ``ground truth'' image contains much less noise than the noisy image and has much better visual quality. Hence, this dataset provides us a good platform for evaluating the image denoising methods. The detailed description on cameras and camera settings is listed in Table \ref{tab6-4}. One can see that in our dataset, the ISO values are more comprehensive than the previous datasets \cite{RENOIR2014,crosschannel2016,dnd2017}.
 
\begin{table*}[t!]
\caption{Cameras and camera settings used in our new dataset.}
\label{tab6-4}
\begin{center}
\small
\renewcommand\arraystretch{1.2}
\begin{tabular*}{1\textwidth}{@{\extracolsep{\fill}}ccccc}
\hline
Camera
&
\# of Scenes
&
Sensor Size (mm)
&
\# of Cropped Regions
&
ISO
\\
\hline
Canon 5D & 10  & $36\times24$  & 29  & 3.2k,6.4k
\\
\hline
Canon 80D & 6  & $22.5\times15$  & 15 & 800,1.6k,3.2k,6.4k,12.8k 
\\
\hline   
Canon 600D & 5 & $22.3\times14.9$  & 11  & 1.6k,3.2k 
\\
\hline   
Nikon D800 & 12 & $35.9\times24$  & 33 & 1.6k,1.8k,3.2k,5k,6.4k 
\\
\hline   
Sony A7II & 7 & $35.8\times23.9$  & 12  & 1.6k,3.2k,6.4k 
\\
\hline
\end{tabular*}
\end{center}
\vspace{-5mm}
\end{table*}

\section{Experiments}

\subsection{Benchmark Datasets} 

To better evaluate the effectiveness of existing image denoising methods, we apply the competing methods on existing datasets \cite{crosschannel2016,dnd2017} and our constructed new dataset. Since the captured real-world noisy images in \cite{RENOIR2014} have clear color differences with the corresponding ``ground truth'' images, we do not evaluate image denoising methods on this dataset.

\textbf{Dataset 1} is provided in \cite{crosschannel2016}, which includes noisy images of 11 static scenes captured by Canon 5D Mark 3, Nikon D600, and Nikon D800 cameras.\ 15 regions of size $512\times512$ were cropped to evaluate different denoising methods.  

\textbf{Dataset 2} is called the Darmstadt Noise Dataset (DND) \cite{dnd2017}, which includes 50 different pairs of images of the same scenes captured by Sony A7R, Olympus E-M10, Sony RX100 IV, and Huawei Nexus 6P.\ The authors cropped 20 bounding boxes of $512\times512$ pixels from each image in the dataset, yielding 1,000 testing crops in total.\ However, the ``ground truth'' images are not open access, and we can only submit the denoising results to the authors' \href{https://noise.visinf.tu-darmstadt.de/}{Project Website} and get the PSNR and SSIM \cite{ssim} results.

\textbf{Dataset 3} is our constructed dataset, which includes noisy images of 40 static scenes captured by Canon 5D Mark II, Canon 80D, Canon 600D, Nikon D800, and Sony A7 II cameras.\ 100 regions of size $512\times512$ were cropped to evaluate different denoising methods.

\subsection{Comparison Methods}

With the proposed dataset, we make a comprehensive evaluation on the state-of-the-art image denoising methods, including CBM3D \cite{cbm3d}, Expected Patch Log Likelihood (EPLL) \cite{epll}, Patch Group Prior based Denoising (PGPD) \cite{pgpd}, Nonlocally Centralized Sparse Representations (NCSR) \cite{ncsr}, Weighted Nuclear Norm Minimization (WNNM) \cite{wnnm}, multi-layer perception (MLP) \cite{mlp}, Cascades of Shrinkage Fileds (CSF) \cite{csf}, Trainable Nonlinear Reactive Diffusion (TNRD) \cite{tnrd}, the residual network based method DnCNN \cite{dncnn}, the ``Noise Clinic'' method \cite{noiseclinic,ncwebsite}, the commercial software Neat Image \cite{neatimage}, and the recently proposed methods external prior guided internal prior learning for image denoising (Guided) \cite{guided}, Multi-channel Weighted Nuclear Norm Minimization (MCWNNM) \cite{mcwnnm}, the Trilateral Weighted Sparse Coding (TWSC) \cite{twsc}. CBM3D is a state-of-the-art color image denoising method, which assumes that the noise is AWGN. The EPLL, PGPD, NCSR, WNNM, MLP, CSF, TNRD, DnCNN are state-of-the-art methods for AWGN noise removal on greyscale images, and we apply these methods on each channel of the realsitic color images.  The ``Noise Clnic'' (NC) is a blind image denoising method while Neat Image (NI) is a set of commercial software for image denoising, which has been embedded into Photoshop and Corel Paint Shop. Besides, the method of DnCNN \cite{dncnn} can also deal with real-world noisy images. the recently proposed methods includes Guided, MCWNNM, and TWSC, which are proposed for real-world noisy image denoising.

\begin{figure*}%[t!]
    \centering
\subfigure{
\begin{minipage}[t]{0.24\textwidth}
\centering
\raisebox{-0.5cm}{\includegraphics[width=1\textwidth]{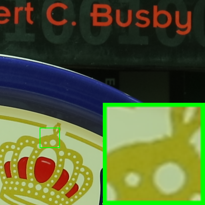}}
{\footnotesize Mean Image}
\end{minipage}
\begin{minipage}[t]{0.24\textwidth}
\centering
\raisebox{-0.5cm}{\includegraphics[width=1\textwidth]{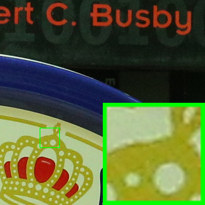}}
{\footnotesize Noisy 37.00/0.9345}
\end{minipage}
\begin{minipage}[t]{0.24\textwidth}
\centering
\raisebox{-0.5cm}{\includegraphics[width=1\textwidth]{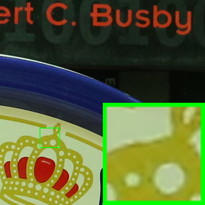}}
{\footnotesize CBM3D 39.72/0.9769}
\end{minipage}
\begin{minipage}[t]{0.24\textwidth}
\centering
\raisebox{-0.5cm}{\includegraphics[width=1\textwidth]{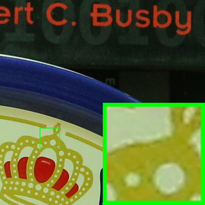}}
{\footnotesize EPLL 37.61/0.9521}
\end{minipage}
}\vspace{-3mm}
\subfigure{
\begin{minipage}[t]{0.24\textwidth}
\centering
\raisebox{-0.5cm}{\includegraphics[width=1\textwidth]{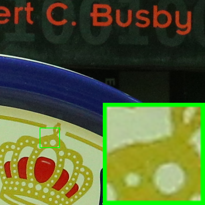}}
{\footnotesize PGPD 37.50/0.9457}
\end{minipage}
\begin{minipage}[t]{0.24\textwidth}
\centering
\raisebox{-0.5cm}{\includegraphics[width=1\textwidth]{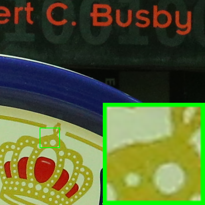}}
{\footnotesize NCSR 37.93/0.9579}
\end{minipage}
\begin{minipage}[t]{0.24\textwidth}
\centering
\raisebox{-0.5cm}{\includegraphics[width=1\textwidth]{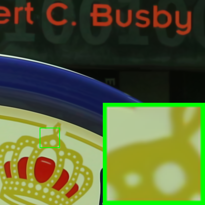}}
{\footnotesize WNNM 37.48/0.9664}
\end{minipage}
\begin{minipage}[t]{0.24\textwidth}
\centering
\raisebox{-0.5cm}{\includegraphics[width=1\textwidth]{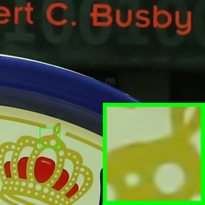}}
{\footnotesize MLP 39.00/0.9695}
\end{minipage}
}\vspace{-3mm}
\subfigure{
\begin{minipage}[t]{0.24\textwidth}
\centering
\raisebox{-0.5cm}{\includegraphics[width=1\textwidth]{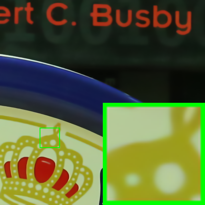}}
{\footnotesize CSF 35.66/0.9425}
\end{minipage}
\begin{minipage}[t]{0.24\textwidth}
\centering
\raisebox{-0.5cm}{\includegraphics[width=1\textwidth]{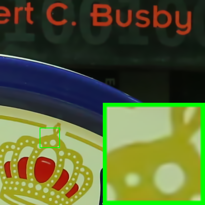}}
{\footnotesize TNRD 39.46/0.9733}
\end{minipage}
\begin{minipage}[t]{0.24\textwidth}
\centering
\raisebox{-0.5cm}{\includegraphics[width=1\textwidth]{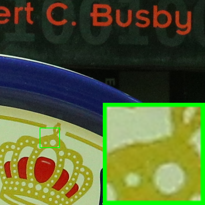}}
{\footnotesize DnCNN 37.26/0.9389}
\end{minipage}
\begin{minipage}[t]{0.24\textwidth}
\centering
\raisebox{-0.5cm}{\includegraphics[width=1\textwidth]{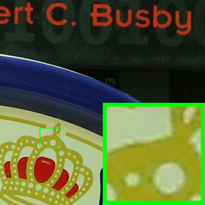}}
{\footnotesize NC 38.76/0.9689}
\end{minipage}
}\vspace{-3mm}
\subfigure{
\begin{minipage}[t]{0.24\textwidth}
\centering
\raisebox{-0.5cm}{\includegraphics[width=1\textwidth]{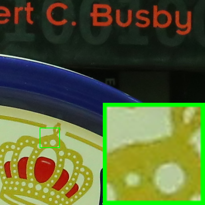}}
{\footnotesize NI 37.68/0.9600}
\end{minipage}
\begin{minipage}[t]{0.24\textwidth}
\centering
\raisebox{-0.5cm}{\includegraphics[width=1\textwidth]{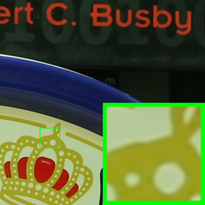}}
{\footnotesize Guided 40.52/0.9804}
\end{minipage}
\begin{minipage}[t]{0.24\textwidth}
\centering
\raisebox{-0.5cm}{\includegraphics[width=1\textwidth]{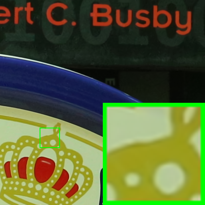}}
{\footnotesize MCWNNM 40.71/0.9775}
\end{minipage}
\begin{minipage}[t]{0.24\textwidth}
\centering
\raisebox{-0.5cm}{\includegraphics[width=1\textwidth]{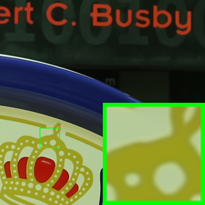}}
{\footnotesize TWSC 40.70/0.9796}
\end{minipage}
}\vspace{-3mm}
    \caption{Denoised images and PSNR (dB)/SSIM results of the real-world noisy image \textsl{Canon 5D Mark 3 ISO 3200 1} \cite{crosschannel2016} by different methods.\ The images are better to be zoomed in on screen.}
    \label{fig6-5}
\end{figure*}

\textbf{Noise level estimation for comparison methods.} For the CBM3D method, the standard deviation of noise on color images should be given as a parameter. For methods of EPLL, PGPD, NCSR,  WNNM, MLP, CSF, and TNRD, the noise level in each color channel should be input. For the DnCNN method, it is trained to deal with noise in a range of levels $0\sim55$. We retrain the models of discriminative denoising methods MLP, CSF, and TNRD (using the released codes by the authors) at different noise levels from $\sigma=5$ to $\sigma=50$ with a gap of 5. The denoising is performed by processing each channel with the model trained at the same (or nearest) noise level. The noise levels ($\sigma_{r}, \sigma_{g}, \sigma_{b}$) in R, G, B channels are estimated via some noise estimation methods \cite{noiselevel,Chen2015ICCV}. In this chapter, we employ the method \cite{noiselevel} to estimate the noise level for each channel of the input color image.

\begin{table*}[t!]
\caption{Average results on PSNR(dB) and SSIM of different denoising algorithms on the 15 cropped images in \textbf{Dataset 1} \cite{crosschannel2016}.}
\footnotesize
\label{tab6-6}
\begin{center}
\renewcommand\arraystretch{1.2}
\begin{tabular*}{1\textwidth}{@{\extracolsep{\fill}}cccccccc}
\hline
Metric
&
{CBM3D}
&
{EPLL}
&
{PGPD}
&
{NCSR}
&
{WNNM}
&
{MLP}
&
{CSF}
\\
\hline
PSNR & 35.19  & 33.66 & 33.69 & 33.46 &  35.77 &  36.46 & 35.33  
\\
\hline
SSIM & 0.8580 & 0.8591  & 0.8591 & 0.8512 & 0.9381 &  0.9436  & 0.9250 
\\
\hline
Metric
&
{TNRD}
&
{DnCNN}
&
{NC}
&
{NI}
&
{Guided}
&
{MCWNNM}
&
{TWSC}
\\
\hline
PSNR  & 36.61 & 33.86 & 36.43 & 35.49 & 37.15 & 37.71 & \textbf{37.81}
\\
\hline
SSIM & 0.9463 & 0.8635 & 0.9364 & 0.9126 & 0.9504 & 0.9542 & \textbf{0.9586}
\\
\hline
\end{tabular*}
\end{center}
\end{table*}

\begin{table*}[t!]
\caption{Average results on PSNR(dB) and SSIM of different denoising algorithms on the 1,000 cropped images in \textbf{Dataset 2} \cite{dnd2017}.}
\footnotesize
\label{tab6-7}
\begin{center}
\renewcommand\arraystretch{1.2}
\begin{tabular*}{1\textwidth}{@{\extracolsep{\fill}}cccccccc}
\hline
Metric
&
{CBM3D}
&
{EPLL}
&
{PGPD}
&
{NCSR}
&
{WNNM}
&
{MLP}
&
{CSF}
\\
\hline
PSNR & 32.14 &  32.65 & 33.12 & 32.81 & 33.28  & 34.02  & 33.87 
\\
\hline
SSIM & 0.7773 & 0.7889 & 0.8002 & 0.7912  & 0.8012  &  0.8201 & 0.8128 
\\
\hline
Metric
&
{TNRD}
&
{DnCNN}
&
{NC}
&
{NI}
&
{Guided}
&
{MCWNNM}
&
{TWSC}
\\
\hline
PSNR & 34.15 & 32.41 & 36.07 & 35.11 & 36.41 & 37.38 &  \textbf{37.94}
\\
\hline
SSIM & 0.8271 & 0.7897 & 0.9013 & 0.8778 & 0.9101 & 0.9294 &  \textbf{0.9403}
\\
\hline
\end{tabular*}
\end{center}
\end{table*}

\begin{figure*}[ht!]
    \centering
\subfigure{
\begin{minipage}[t]{0.19\textwidth}
\centering
\raisebox{-0.5cm}{\includegraphics[width=1\textwidth]{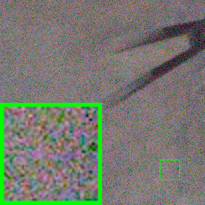}}
{\footnotesize Noisy }
\end{minipage}
\begin{minipage}[t]{0.19\textwidth}
\centering
\raisebox{-0.5cm}{\includegraphics[width=1\textwidth]{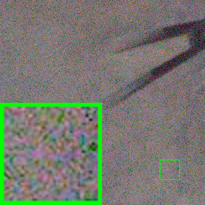}}
{\footnotesize CBM3D}
\end{minipage}
\begin{minipage}[t]{0.19\textwidth}
\centering
\raisebox{-0.5cm}{\includegraphics[width=1\textwidth]{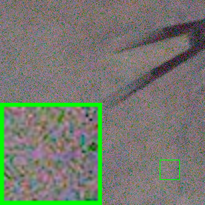}}
{\footnotesize EPLL}
\end{minipage}
\begin{minipage}[t]{0.19\textwidth}
\centering
\raisebox{-0.5cm}{\includegraphics[width=1\textwidth]{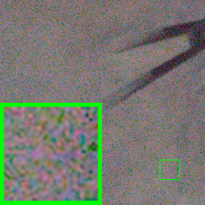}}
{\footnotesize PGPD}
\end{minipage}
\begin{minipage}[t]{0.19\textwidth}
\centering
\raisebox{-0.5cm}{\includegraphics[width=1\textwidth]{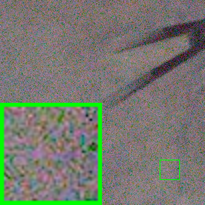}}
{\footnotesize NCSR}
\end{minipage}
}\vspace{-3mm}
\subfigure{
\begin{minipage}[t]{0.19\textwidth}
\centering
\raisebox{-0.5cm}{\includegraphics[width=1\textwidth]{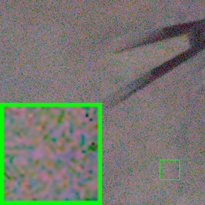}}
{\footnotesize WNNM}
\end{minipage}
\begin{minipage}[t]{0.19\textwidth}
\centering
\raisebox{-0.5cm}{\includegraphics[width=1\textwidth]{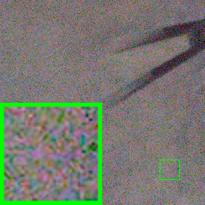}}
{\footnotesize MLP}
\end{minipage}
\begin{minipage}[t]{0.19\textwidth}
\centering
\raisebox{-0.5cm}{\includegraphics[width=1\textwidth]{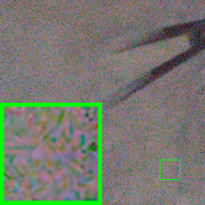}}
{\footnotesize CSF}
\end{minipage}
\begin{minipage}[t]{0.19\textwidth}
\centering
\raisebox{-0.5cm}{\includegraphics[width=1\textwidth]{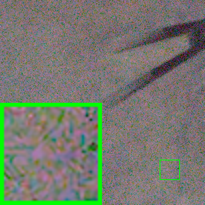}}
{\footnotesize TNRD}
\end{minipage}
\begin{minipage}[t]{0.19\textwidth}
\centering
\raisebox{-0.5cm}{\includegraphics[width=1\textwidth]{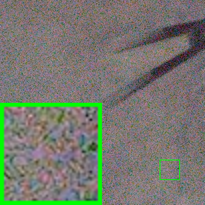}}
{\footnotesize DnCNN}
\end{minipage}
}\vspace{-3mm}
\subfigure{
\begin{minipage}[t]{0.19\textwidth}
\centering
\raisebox{-0.5cm}{\includegraphics[width=1\textwidth]{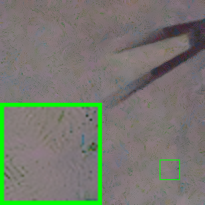}}
{\footnotesize NC}
\end{minipage}
\begin{minipage}[t]{0.19\textwidth}
\centering
\raisebox{-0.5cm}{\includegraphics[width=1\textwidth]{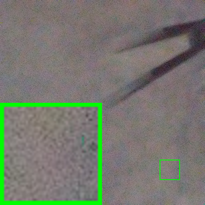}}
{\footnotesize NI}
\end{minipage}
\begin{minipage}[t]{0.19\textwidth}
\centering
\raisebox{-0.5cm}{\includegraphics[width=1\textwidth]{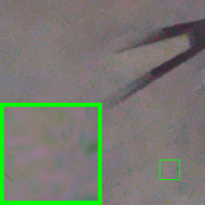}}
{\footnotesize Guided}
\end{minipage}
\begin{minipage}[t]{0.19\textwidth}
\centering
\raisebox{-0.5cm}{\includegraphics[width=1\textwidth]{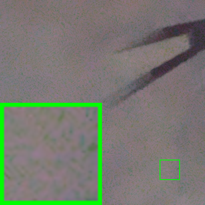}}
{\footnotesize MCWNNM}
\end{minipage}
\begin{minipage}[t]{0.19\textwidth}
\centering
\raisebox{-0.5cm}{\includegraphics[width=1\textwidth]{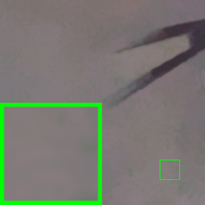}}
{\footnotesize TWSC}
\end{minipage}
}
    \caption{Denoised images of the real-world noisy image \textsl{0001\_1} \cite{dnd2017} by different methods.\ The images are better to be zoomed in on screen.}
    \label{fig6-6}
\end{figure*}

\subsection{Results and Discussion}

\textbf{Results on Dataset 1}. The average PSNR and SSIM results on the 15 cropped images by competing methods are listed in Table \ref{tab6-6}.\ One can see that the recently proposed Guided, MCWNNM, and  TWSC methods perform much better than the other competing methods.\ Figure\ \ref{fig6-5} shows the denoised images of a scene captured by Canon 5D Mark 3 at ISO = 3200.\ One can see that TWSC results in not only higher PSNR and SSIM measurements, but also much better visual quality than other methods.

\textbf{Results on Dataset 2}. In Table \ref{tab6-7}, we list the average PSNR and SSIM results of the competing methods on the 1,000 cropped images in the DND dataset \cite{dnd2017}.\ We can see that on this dataset the Guided, MCWNNM, and  TWSC methods achieve much better performance than the other competing methods.\ Note that the ``ground truth'' images of this dataset have not been released, but one can submit the denoised images to the project website and get the PSNR and SSIM results.\ Figure\ \ref{fig6-6} shows the denoised images of a scene captured by a Nexus 6P camera.\ One can see that the TWSC method results in better visual quality than the other denoising methods.

\begin{figure*}%[ht!]
    \centering
\subfigure{
\begin{minipage}[t]{0.24\textwidth}
\centering
\raisebox{-0.5cm}{\includegraphics[width=1\textwidth]{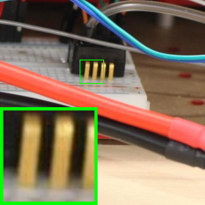}}
{\footnotesize Mean Image}
\end{minipage}
\begin{minipage}[t]{0.24\textwidth}
\centering
\raisebox{-0.5cm}{\includegraphics[width=1\textwidth]{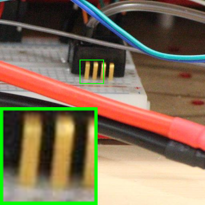}}
{\footnotesize Noisy 33.13/0.9091}
\end{minipage}
\begin{minipage}[t]{0.24\textwidth}
\centering
\raisebox{-0.5cm}{\includegraphics[width=1\textwidth]{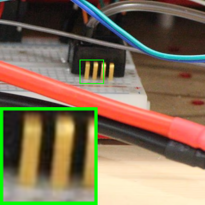}}
{\footnotesize CBM3D 33.87/0.9412}
\end{minipage}
\begin{minipage}[t]{0.24\textwidth}
\centering
\raisebox{-0.5cm}{\includegraphics[width=1\textwidth]{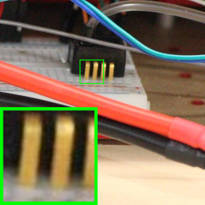}}
{\footnotesize EPLL 33.21/0.9142}
\end{minipage}
}\vspace{-3mm}
\subfigure{
\begin{minipage}[t]{0.24\textwidth}
\centering
\raisebox{-0.5cm}{\includegraphics[width=1\textwidth]{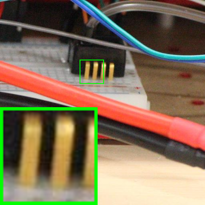}}
{\footnotesize PGPD 33.18/0.9116}
\end{minipage}
\begin{minipage}[t]{0.24\textwidth}
\centering
\raisebox{-0.5cm}{\includegraphics[width=1\textwidth]{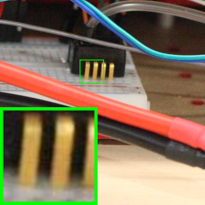}}
{\footnotesize NCSR 33.28/0.9182}
\end{minipage}
\begin{minipage}[t]{0.24\textwidth}
\centering
\raisebox{-0.5cm}{\includegraphics[width=1\textwidth]{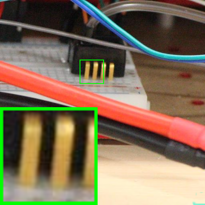}}
{\footnotesize WNNM 32.99/0.9179}
\end{minipage}
\begin{minipage}[t]{0.24\textwidth}
\centering
\raisebox{-0.5cm}{\includegraphics[width=1\textwidth]{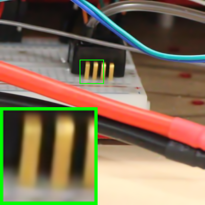}}
{\footnotesize MLP 34.33/0.9584}
\end{minipage}
}\vspace{-3mm}
\subfigure{
\begin{minipage}[t]{0.24\textwidth}
\centering
\raisebox{-0.5cm}{\includegraphics[width=1\textwidth]{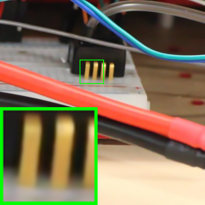}}
{\footnotesize CSF 34.13/0.9546}
\end{minipage}
\begin{minipage}[t]{0.24\textwidth}
\centering
\raisebox{-0.5cm}{\includegraphics[width=1\textwidth]{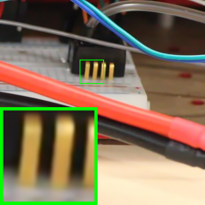}}
{\footnotesize TNRD 34.37/0.9594}
\end{minipage}
\begin{minipage}[t]{0.24\textwidth}
\centering
\raisebox{-0.5cm}{\includegraphics[width=1\textwidth]{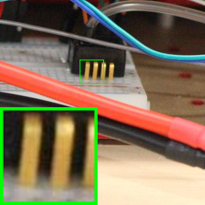}}
{\footnotesize DnCNN 33.26/0.9137}
\end{minipage}
\begin{minipage}[t]{0.24\textwidth}
\centering
\raisebox{-0.5cm}{\includegraphics[width=1\textwidth]{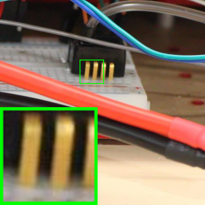}}
{\footnotesize NC 33.33/0.9357}
\end{minipage}
}\vspace{-3mm}
\subfigure{
\begin{minipage}[t]{0.24\textwidth}
\centering
\raisebox{-0.5cm}{\includegraphics[width=1\textwidth]{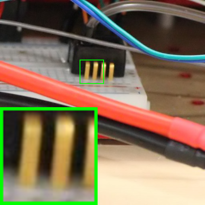}}
{\footnotesize NI 34.16/0.9500}
\end{minipage}
\begin{minipage}[t]{0.24\textwidth}
\centering
\raisebox{-0.5cm}{\includegraphics[width=1\textwidth]{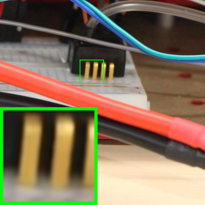}}
{\footnotesize Guided 34.68/0.9622}
\end{minipage}
\begin{minipage}[t]{0.24\textwidth}
\centering
\raisebox{-0.5cm}{\includegraphics[width=1\textwidth]{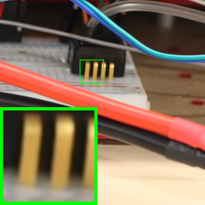}}
{\footnotesize MCWNNM 34.70/0.9629}
\end{minipage}
\begin{minipage}[t]{0.24\textwidth}
\centering
\raisebox{-0.5cm}{\includegraphics[width=1\textwidth]{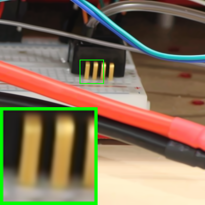}}
{\footnotesize TWSC 34.72/0.9640}
\end{minipage}
}\vspace{-3mm}
\caption{Denoised images and PSNR (dB)/SSIM results of the real-world noisy image \textsl{Canon5D\_2.5\_160\_6400\_circuit\_3} in our new dataset by different methods.\ The images are better to be zoomed in on screen.}
    \label{fig6-7}
\end{figure*}

\textbf{Results on Dataset 3}.
The PSNR and SSIM \cite{ssim} results on 100 images of the cropped regions are listed in Table \ref{tab6-8}. We can see that the traditional methods proposed for AWGN are no longer effective enough for the real-world noisy images. The discriminative methods achieve slightly better performance than the traditional methods, while still being inferior to the methods designed for real-world nosiy images. The recently proposed Guided, MCWNNM, and TWSC methods work much better than previous methods. Some visual comparisons are given in Figure \ref{fig6-7}, from which one can see that the TWSC method removes most of the noise while maintaining the details. 

\begin{figure*}%[ht!]
\vspace{-6mm}
    \centering
\subfigure{
\begin{minipage}[t]{0.48\textwidth}
\centering
\raisebox{-0.5cm}{\includegraphics[width=1\textwidth]{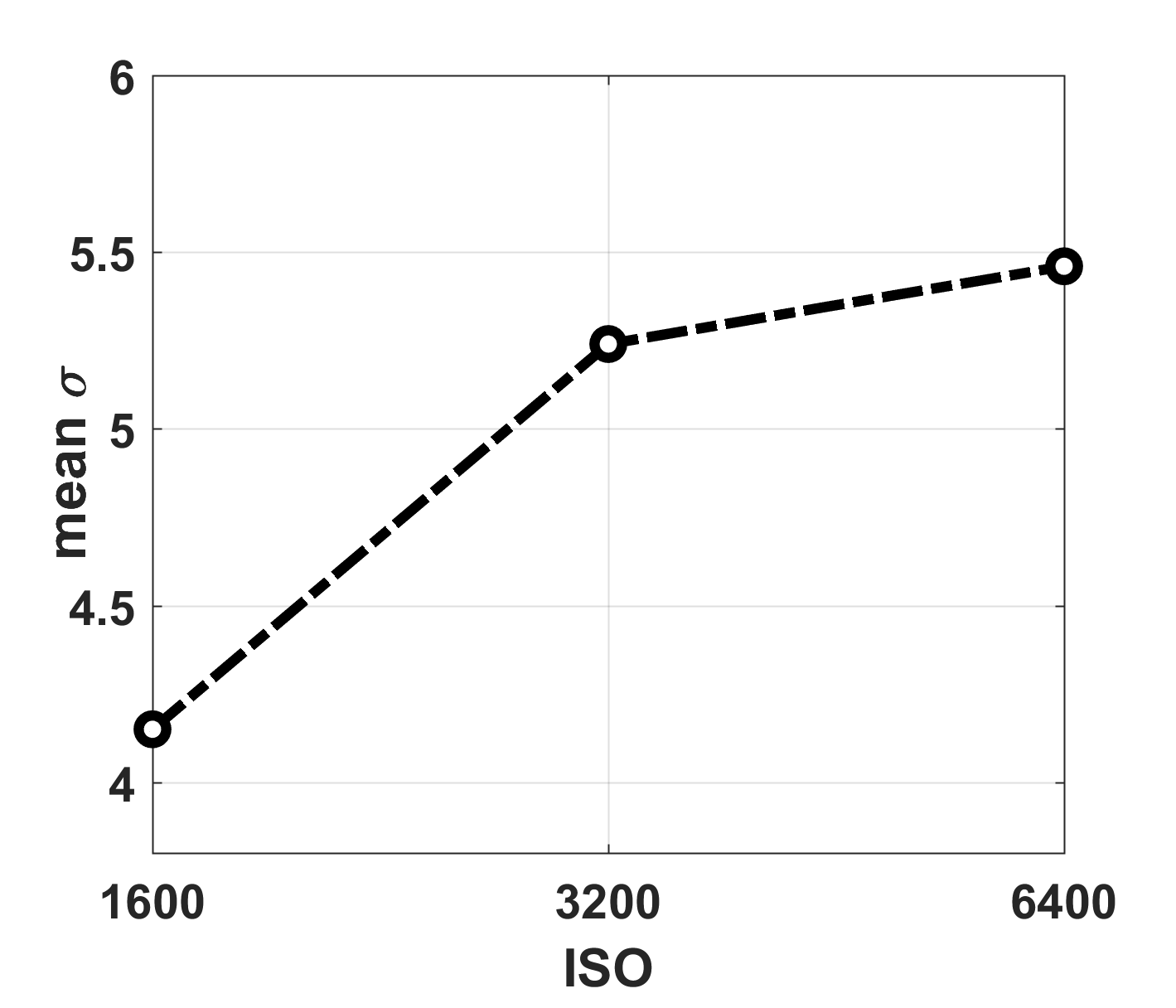}}
\end{minipage}
\begin{minipage}[t]{0.48\textwidth}
\centering
\raisebox{-0.5cm}{\includegraphics[width=1\textwidth]{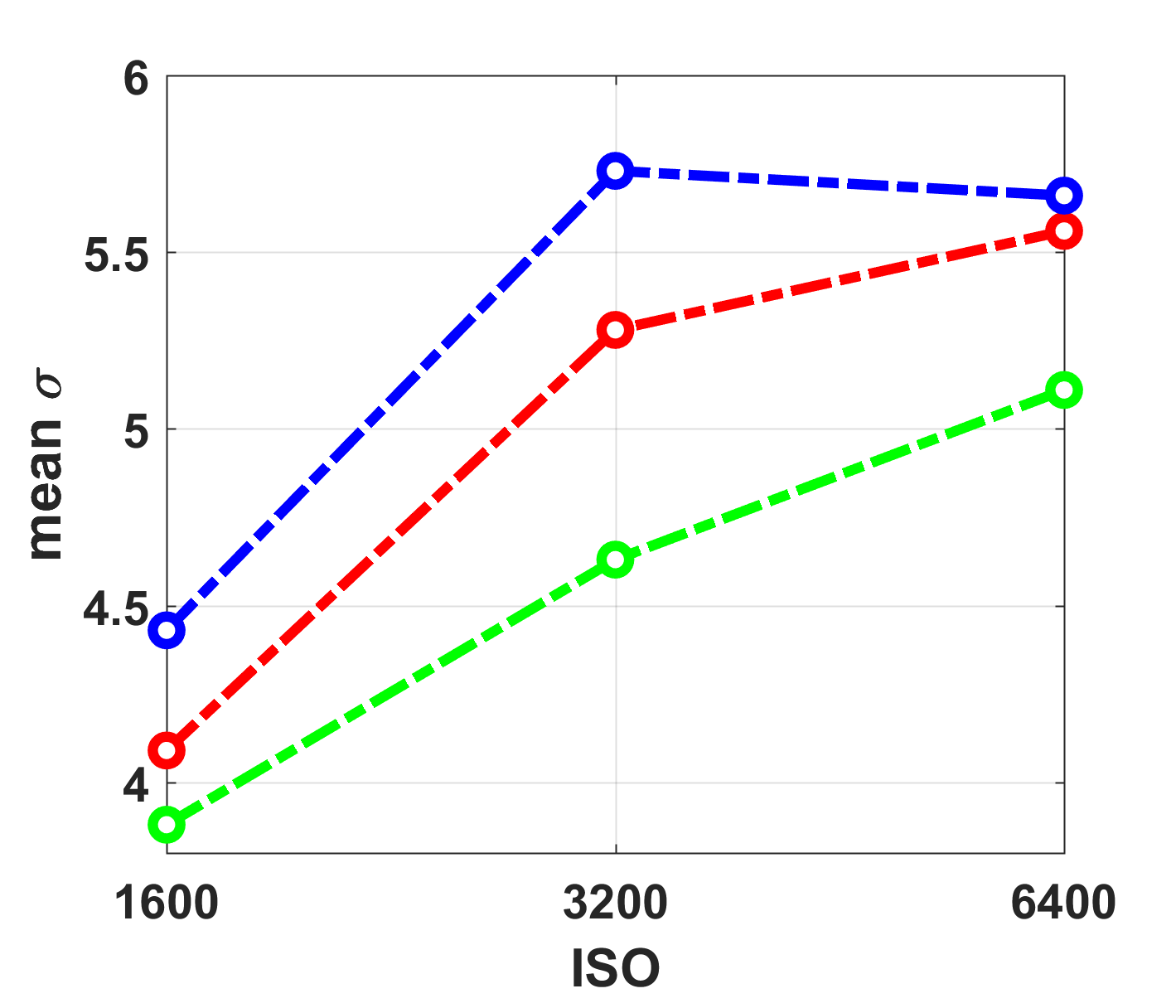}}
\end{minipage}
}\vspace{-1mm}
\caption{The mean noise levels ($\sigma$) with respect to different ISO values (left) and Red, Green, and Blue channels (right) on our real-world noisy image dataset.}
\label{fig6-8}
\end{figure*}

The realistic noise is not AWGN, and this point can be valid in our new dataset. In Figure \ref{fig6-8}, we compute the mean noise levels ($\sigma$) with respect to different ISO values (left) and Red, Green, and Blue channels (right) on our real-world noisy image dataset. One can see that, with the increasing of the ISO values, the noise levels will be increased. This trend is also true for each channel of the color images. One possible correlation is that the noise level in Green channel would be lower than the other two channels, i.e., $\sigma_{b},\sigma_{r}\ge\sigma_{g}$.

\begin{table*}[t!]
\caption{Average results on PSNR(dB) and SSIM of different denoising algorithms on the 100 cropped images in our new dataset (\textbf{Dataset 3}).}
\footnotesize
\label{tab6-8}
\begin{center}
\renewcommand\arraystretch{1.2}
\begin{tabular*}{1\textwidth}{@{\extracolsep{\fill}}cccccccc}
\hline
Metric
&
{CBM3D}
&
{EPLL}
&
PGPD
&
{NCSR}
&
{WNNM}
&
{MLP}
&
{CSF}
\\
\hline
PSNR & 37.40 & 36.17 & 36.18 & 36.40 & 36.59 & 38.07 & 37.71 
\\
\hline
SSIM & 0.9526 & 0.9216 & 0.9206 & 0.9290 & 0.9247 & 0.9615 & 0.9571 
\\
\hline
Metric
&
{TNRD}
&
{DnCNN}
&
{NC}
&
{NI}
&
{Guided}
&
{MCWNNM}
&
{TWSC}
\\
\hline
PSNR & 38.17 & 36.08 & 36.92  &  37.77 & 38.35 & 38.51 & \textbf{38.60}
\\
\hline
SSIM & 0.9640 & 0.9161 & 0.9449  & 0.9570  & 0.9669 & 0.9671 & \textbf{0.9685}
\\
\hline
\end{tabular*}
\end{center}
\end{table*}

\textbf{Discussion}. 
The experimental results on the three datasets demonstrate that:
\begin{itemize}
\item The methods CBM3D, EPLL, PGPD, NCSR, WNNM, MLP, CSF, TNRD, and DnCNN which are designed for AWGN achieve lower PSNR and SSIM compared with the methods developed for real-world noisy image denoising;

\item The denoising methods EPLL, PGPD, NCSR, and WNNM which are designed for grey scale images would generate much artifacts since they process each channel of the RGB image individually \cite{srcolor}. They cannot deal with the images which have different noise statistics in different channels as well as different local patches. Hence, these methods may fail to process the real-world noisy images which have complex noise statistics;

\item The discriminative learning based methods MLP, CSF, TNRD, and DnCNN are trained on paired clean and noisy images. These methods largely depends on the training dataset, and would achieve inferior performance on images whose noise has different statistics from those in the training images. Besides, discriminative methods will also be sensitive to the resolution (DPI) of the images in the training set;

\item The performance of recently proposed denoising methods, i.e., Guided, MCWNNM, and TWSC, on datasets \cite{crosschannel2016,dnd2017} are much better than those of previous denoising methods. As we can see from Tables \ref{tab6-6} and \ref{tab6-7}, the highest PSNR of these methods (TWSC) and the highest PSNR of the methods proposed for AWGN removal (TNRD on Dataset 1 and NC on Dataset 2) have a difference of over 1.2dB. However, on our new dataset, as we can see from Tables \ref{tab6-8}, the highest PSNR of these methods (TWSC) and the highest PSNR of previous methods (TNRD) only have a difference of around 0.4dB. This indicates that on our new dataset, the recently proposed methods such as Guided, MCWNNM, and TWSC do not show significant advantages over the previous methods such as TNRD. This is because our dataset is more comprehensive in the scene contents and have more camera settings. This also shows that our dataset is more challenging than previous datasets, and new real-world image denoising methods are needed.
  
\end{itemize}

\section{Conclusion}

To evaluate the existing denoising methods on real photographs and promote new methods for removing real-world noise, we constructed a novel dataset which contains comprehensive real-world noisy images of different natural scenes. These images were captured by different cameras under different camera settings. Each scene was shot 500 times in a short time. We first selected the images without misalignment, then deleted the images which do not have consistant lumiance with the baseline image. Since the captured images are very large in size, we cropped smaller regions of size $512\times512$ to evaluate the existing image denoising methods and the methods we proposed in  previous sections. We evaluated the different denoising methods on the new dataset and previous datasets. The results demonstrated that the proposed methods are more robust than other competing methods, and the newly proposed dataset is more challenging. We will make the constructed dataset of real photographs publicly available for researchers to investigate new real-world image denoising methods.

{
\small
\bibliographystyle{unsrt}
\bibliography{egbib}
}

\end{document}